%% file: main.tex
\documentclass[10pt,twocolumn,letterpaper]{article}

\usepackage[pagenumbers]{cvpr} 

\usepackage{graphicx}
\usepackage{amsmath}
\usepackage{amssymb}
\usepackage{booktabs}
\usepackage{xcolor}
\usepackage{multirow} 
\usepackage{etoc}
\usepackage{minitoc}
\usepackage[accsupp]{axessibility}

\usepackage{algorithm}
\usepackage{algorithmicx}

\usepackage{bm}

\usepackage{xcolor}
\PassOptionsToPackage{hyphens}{url}\usepackage[pagebackref=true,breaklinks=true,letterpaper=true,colorlinks,
  citecolor=citecolor,bookmarks=false]{hyperref}
\definecolor{citecolor}{RGB}{34,139,34}

\usepackage[capitalize]{cleveref}
\crefname{section}{Sec.}{Secs.}
\Crefname{section}{Section}{Sections}
\Crefname{table}{Table}{Tables}
\crefname{table}{Tab.}{Tabs.}

\def\cvprPaperID{} 
\def\confName{CVPR}
\def\confYear{2023}

\def\themodel{NeuralEditor\xspace}

\begin{document}

\title{\themodel: Editing Neural Radiance Fields via Manipulating Point Clouds}

\author{Jun-Kun Chen$^{1}$$^\dagger$ \qquad Jipeng Lyu$^{2}$$^\dagger$ \qquad Yu-Xiong Wang$^{1}$ \vspace{0.1em} \\ 
    $^1$University of Illinois at Urbana-Champaign \qquad $^2$Peking University \qquad $^\dagger$Equal Contribution \vspace{0.1em}\\
    {\tt \hspace{0mm}\{junkun3, yxw\}@illinois.edu}\qquad{\tt lvjipeng@pku.edu.cn}
}

\input{0.1-teaser.tex}

\input{0.2-abstract.tex}

\input{1-intro.tex}
\input{2-related.tex}

\input{3-method.tex}
\input{4-exp}
\input{5-conclusion.tex}

{\small
\bibliographystyle{ieee_fullname}
\bibliography{egbib}
}

\newpage
\input{6-appendix}

\end{document}

%% file: 0.1-teaser.tex
\twocolumn[{
\maketitle
\renewcommand\twocolumn[1][]{#1}
    \centering
    \vspace{-3.0mm}
    \includegraphics[width=1.0\linewidth]{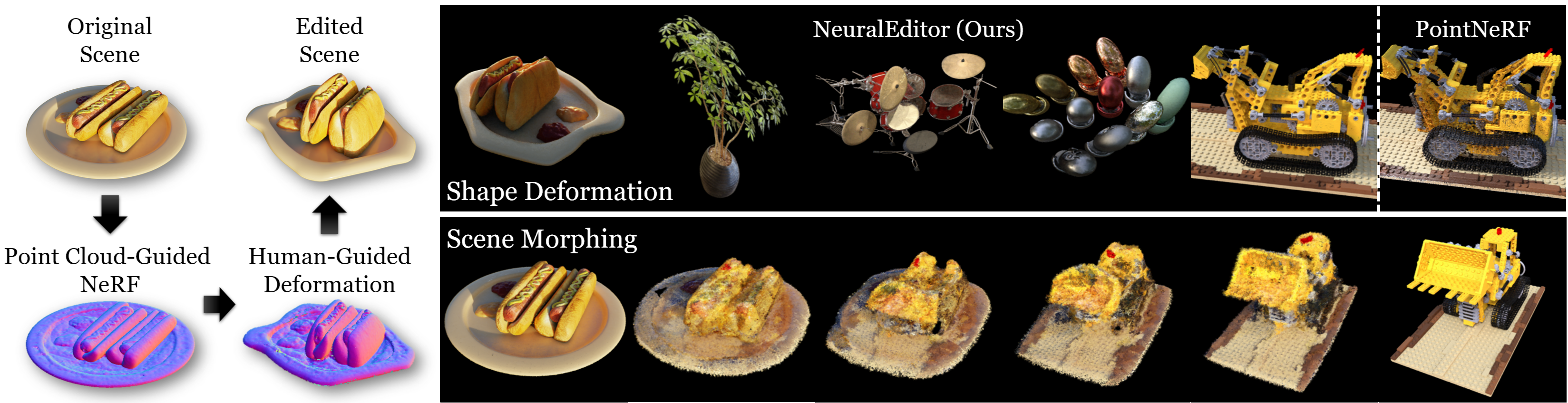}
    \captionof{figure}{Our {\themodel} offers {\em native} support for general and flexible shape editing of neural radiance fields via manipulating point clouds. By generating a precise point cloud of the scene with a novel point cloud-guided NeRF model, our \themodel produces high-fidelity rendering results in both shape deformation and more challenging scene morphing tasks. } 
\label{fig:teaser}
    \vspace{3mm}
}]

%% file: 0.2-abstract.tex
\begin{abstract}

\vspace{-0.8em}

This paper proposes {\themodel} that enables neural radiance fields (NeRFs) natively editable for general shape editing tasks. Despite their impressive results on novel-view synthesis, it remains a fundamental challenge for NeRFs to edit the shape of the scene. Our key insight is to exploit the explicit point cloud representation as the underlying structure to construct NeRFs, inspired by the intuitive interpretation of NeRF rendering as a process that projects or ``plots'' the associated 3D point cloud to a 2D image plane. To this end, {\themodel} introduces a novel rendering scheme based on deterministic integration within K-D tree-guided density-adaptive voxels, which produces both high-quality rendering results and precise point clouds through optimization. \themodel then performs shape editing via mapping associated points between point clouds.
Extensive evaluation shows that \themodel achieves state-of-the-art performance in both shape deformation and scene morphing tasks. Notably, {\themodel} supports both zero-shot inference and further fine-tuning over the edited scene. {Our code, benchmark, and demo video are available at \href{https://immortalco.github.io/NeuralEditor/}{\textit{immortalco.github.io/NeuralEditor}}.}

\vspace{-0.7em}

\end{abstract}

%% file: 1-intro.tex
\section{Introduction}
\label{sec:intro}
\label{sec:intro-v3}

Perhaps the most memorable shot of the film {\em Transformers}, {\em Optimus Prime} is seamlessly transformed between a humanoid and a Peterbilt truck -- such free-form editing of 3D objects and scenes is a fundamental task in 3D computer vision and computer graphics, directly impacting applications such as visual simulation, movie, and game industries. In these applications, often we are required to manipulate a scene or objects in the scene by editing or modifying its shape, color, lighting condition, \etc, and generate visually-faithful rendering results on the edited scene efficiently. Among the various editing operations, shape editing has received continued attention but remains challenging, where the scene is deformed in a human-guided way, while all of its visual attributes (\eg, shape, color, brightness, and lighting condition) are supposed to be natural and consistent with the ambient environment.

State-of-the-art rendering models are based on implicit neural representations, as exemplified by neural radiance field (NeRF)~\cite{nerf} and its variants~\cite{mipnerf,refnerf,ibrnet,plenoct,plenoxels}. Despite their impressive novel-view synthesis results, most of the NeRF models substantially lack the ability for users to adjust, edit, or modify the shape of scene objects. On the other hand, shape editing operations can be natively applied to explicit 3D representations such as point clouds and meshes.

Inspired by this, we propose {\em \themodel} -- a general and flexible approach to editing neural radiance fields via manipulating point clouds (Fig.~\ref{fig:teaser}). Our {\em key insight} is to benefit from the best of both worlds: the superiority in rendering performance from implicit neural representation {\em combined with} the ease of editing from explicit point cloud representation. {\themodel} enables us to perform a wide spectrum of shape editing operations in a consistent way.

Such introduction of point clouds into NeRF for general shape editing is rooted in our interpretation of {\em NeRF rendering as a process that projects or ``plots'' the associated 3D point cloud to a 2D image plane}. Conceptually, with a dense enough point cloud where each point has an opacity and its color is defined as a function of viewing direction, directly plotting the point cloud would achieve similar visual effects (\ie, transparency and view-dependent colors) that are rendered by NeRF. This {\em intrinsic integration} between NeRF and point clouds underscores the advantage of our \themodel over existing mesh-based NeRF editing methods such as NeRF-Editing~\cite{nerfediting}, Deforming-NeRF~\cite{deformingnerf}, and CageNeRF~\cite{cagenerf}, where the process of constructing and optimizing the mesh is separated from the NeRF modeling, making them time-consuming. More importantly, with the point cloud constructed for a scene, the shape editing can be natively defined as and easily solved by just moving each point into the new, edited position and re-plotting the point cloud. Therefore, our approach supports more general scene editing operations which are difficult to achieve via mesh-guided space deformation.

The key component in our \themodel lies in a point cloud-guided NeRF model that natively supports general shape editing operations. While the recent method PointNeRF~\cite{pointnerf} has demonstrated improved novel-view synthesis capability based on point clouds, it is not supportive to shape editing. Our idea then is to exploit the underlying point cloud in ways of not only optimizing its structure and features (\eg, adaptive voxels) for rendering, but also extracting additional useful attributes (\eg, normal vectors) to guide the editing process. To this end, we introduce K-D trees~\cite{kdtree} to construct density-adaptive voxels for efficient and stable rendering, together with a novel deterministic integration strategy. Moreover, we model the color with the Phong reflection~\cite{phong} to decompose the specular color and better represent the scene geometry.

With a much more precise point cloud attributed to these improvements, our \themodel achieves high-fidelity rendering results on deformed scenes compared with prior work as shown in Fig.~\ref{fig:teaser}, even in a {\em zero-shot} inference manner without additional training. Through fast fine-tuning, the visual quality of the deformed scene is further enhanced, almost perfectly consistent with the surrounding lighting condition. In addition, under the guidance of a point cloud diffusion model~\cite{clouddiffusion}, \themodel can be naturally extended for smooth {\em scene morphing} across multiple scenes, which is difficult for existing NeRF editing work.

\textbf{Our contributions} are {four}-fold. (1) We introduce \themodel, a flexible and versatile approach that makes neural radiance fields editable through manipulating point clouds. (2) We propose a point cloud-guided NeRF model based on K-D trees and deterministic integration, which produces precise point clouds and supports general scene editing. {(3) Due to the lack of publicly available benchmarks for shape editing, we construct and release a reproducible benchmark that promotes future research on shape editing.} (4) We investigate a wide range of shape editing tasks, covering both shape deformation (as studied in existing NeRF editing work) and challenging scene morphing (a novel task addressed here). \themodel achieves state-of-the-art performance on all shape editing tasks in a unified framework, without extra information or supervision. 

%% file: 2-related.tex
\section{Related Work}
\label{sec:related}

\textbf{Neural Scene Representation.} Traditional methods model scenes with explicit~\cite{tradexp0,tradexp1,tradexp2,tradexp3,tradexp5,tradexp6} or implicit~\cite{tradimp1,tradimp2,tradimp3,tradimp4,nex} 3D geometric or shape representations. Initiated by NeRF~\cite{nerf}, leveraging implicit neural networks to represent scenes and perform novel-view synthesis has become a fast-developing field in 3D vision~\cite{nerfsurvey1,nerfsurvey2}. While most of the follow-up work focuses on improving aspects such as the rendering realism~\cite{mipnerf,nerfren,refnerf,neuralsparse}, efficiency~\cite{plenoct,diver,plenoxels}, and cross-scene generalization~\cite{ibrnet,pixelnerf,pointnerf,mvsnerf}, the scene editing capability is substantially missing in the NeRF family which we address in this paper. In addition, we exploit K-D tree-guided point clouds as the underlying structure, different from other NeRF variants based on octrees~\cite{plenoct,plenoxels,neuralsparse} or plain voxels~\cite{diver}.

\textbf{Point-Based NeRFs.} Recently, using point clouds to build a NeRF model has shown better encoding of scene shape and improved rendering performance, as represented by PointNeRF~\cite{pointnerf}. PointNeRF proposes a point initialization network to produce the initial point cloud together with the point features, which is further optimized by a pruning and growing strategy. While both PointNeRF and our {\themodel} employ point clouds as the underlying structure, {\themodel} better exploits useful information within the point clouds: PointNeRF only directly uses the locations of points; by contrast, {\themodel} {\em considers the point cloud more as a geometrical shape} and extracts relevant information like normal vectors, which plays an important role in rendering and shape editing. Importantly, our approach is designed to support scene editing, in contrast to PointNeRF.

\textbf{Scene Editing via NeRFs.} Different types of scene editing have been studied under NeRFs. EditNeRF~\cite{editnerf}, ObjectNeRF~\cite{objectnerf}, and DistillNeRF~\cite{distillnerf} perform simple shape and color editing for objects specified with human-input scribble, pixel, segment, language, \etc NeuPhysics~\cite{neuphysics} edits a dynamic scene via physics parameters. \mbox{CCNeRF}~\cite{ccnerf} proposes an explicit NeRF representation with tensor rank decomposition to support scene composition. INSP-Net~\cite{insp} considers filter editing like denoising. Such work cannot address 3D shape editing and only supports simple editing operations, like object selection, similarity transformation, or limited shape deformation.

\textbf{3D Shape Editing.} Traditional representation methods support keypoint-based shape editing~\cite{ske1,ske2,cage0,cage1,cage2,cage3,cage4} with meshes~\cite{tradeditsurvey,cagesurvey}, which cannot be directly applied to implicit representations used by NeRF. Existing NeRF editing work primarily studies a particular shape editing task, mesh deformation, and addresses it in a common paradigm~\cite{nerfediting,cagenerf,deformingnerf}: A mesh of the scene is first constructed by either exporting it from a trained NeRF with the Marching Cubes algorithm~\cite{marchingcubes}, or optimizing close-to-surface cages along with training. After the user deforms the mesh, the deformed scene is rendered by deforming the space and bending the viewing rays in the original scene with the trained NeRF. Doing so requires extra efforts to convert {\em implicit} scene representation to {\em explicit} mesh, which might not be precise enough, and only supports {\em continuous} shape editing that can be converted to space deformation. On the contrary, our \themodel directly maintains and utilizes alternative explicit scene representation -- {\em the point cloud which is intrinsically integrated with NeRF}, making \themodel require no extra efforts and support more general shape editing tasks like scene morphing. \themodel supports both zero-shot inference and further fine-tuning over the edited scene, while prior work cannot.

%% file: 3-method.tex
\input{3.1-nerf.tex}

\input{3.2-edit.tex}

%% file: 3.1-nerf.tex
\section{\themodel: Point Cloud-Guided NeRF}
\label{sec:method-nerf}

We propose a novel point cloud-guided NeRF model, {\themodel} -- it not only achieves realistic rendering results in the novel-view synthesis task, but also produces a point cloud that precisely describes the shape of the scene, thus facilitating general shape editing tasks. As illustrated in Fig.~\ref{fig:kd-voxel}, we leverage the K-D trees~\cite{kdtree} to construct {\em density-adaptive} voxels (which also naturally enable us to skip empty spaces), and introduce {\em deterministic} spline integration for rendering. We use the Phong reflection to model the color along with the normal vectors obtained from the underlying point cloud. With our enhanced point cloud optimization, \themodel obtains much more precise underlying point clouds, compared with noisy and imprecise outputs of state-of-the-art PointNeRF~\cite{pointnerf} (as shown in Sec.~\ref{sec:exp}).

\subsection{K-D Tree-Guided Voxels}

To render with points, we construct multi-scale density-adaptive voxels based on K-D trees~\cite{kdtree}, namely, {\em K-D voxels}. K-D trees are a data structure constructed on $K$-dimensional points, where $K=3$ for 3D points. As a special decision tree, K-D tree's each node divides the point set into two equal-sized parts with axis-parallel criterion.

\begin{figure}[t!]
\centering
\centerline{\includegraphics[width=.96\columnwidth]{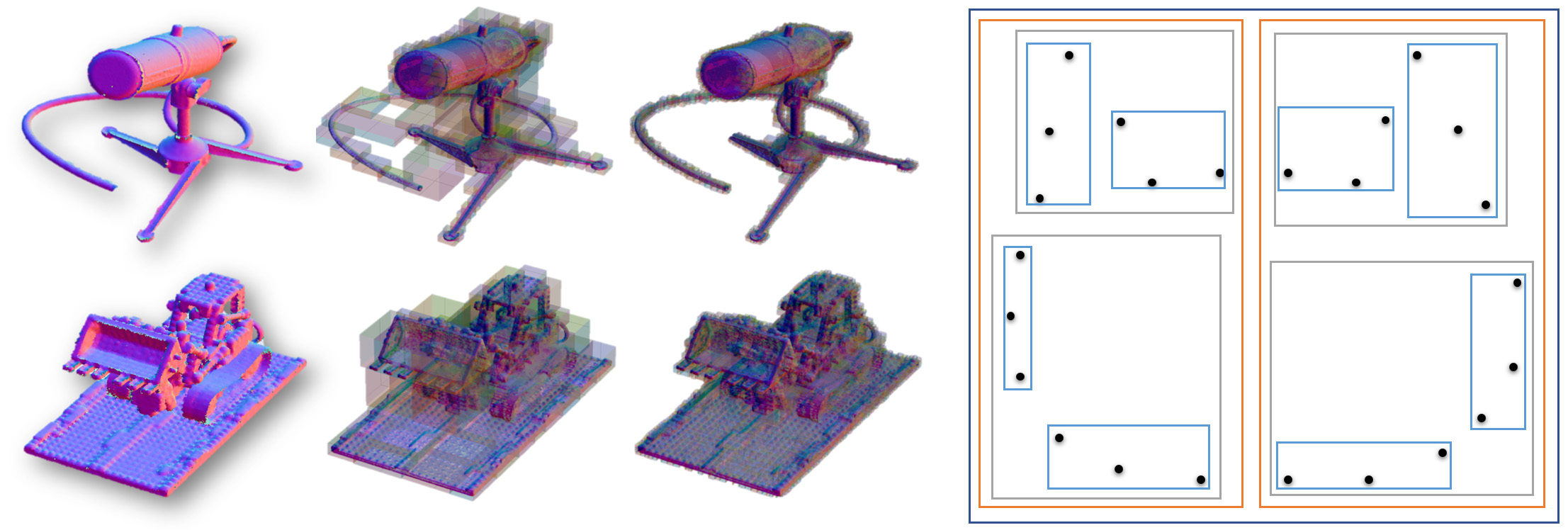}}
\vspace{-3mm}
\caption{\textbf{Our K-D Voxels.} \textbf{Column 1:} original point clouds constructed from two scenes, colored with normal vector directions. \textbf{Column 2:} upper-level voxels which coarsely represent the shape. \textbf{Column 3:} lower-level voxels which tightly cover the shape. {\textbf{Right:} Visualization of K-D voxels on a 2D point cloud. Each color represents boxes of nodes on each level of the K-D tree. Lower-level boxes containing fewer points cover the shape more tightly, and vice versa for higher-level boxes.} }
\label{fig:kd-voxel}
\vspace{-6mm}
\end{figure}

For each K-D tree's node, we compute its bounding box by taking the minimum and maximum $x,y,z$ coordinates in its subtree and with proper padding margins. As we divide the points in a top-down manner in one of the $x,y,z$ directions, in each layer of the K-D tree, different nodes' bounding boxes are {\em mutually exclusive}. Therefore, the bounding boxes can be regarded as voxels. As boxes in the upper layers contain more points (larger voxels), while those in the lower layers contain fewer points (smaller voxels), we natively obtain a {\em multi-scale} voxel construction from one K-D tree. As shown in Fig.~\ref{fig:kd-voxel}, voxels from the large to small scales represent the shape of the scene from coarse to fine.

\subsection{Rendering Over K-D Voxels}
We now introduce a rendering scheme that exploits \mbox{K-D} voxels to perform all the sub-procedures associated with rendering in a unified way. This scheme enables us to render more naturally, efficiently, and even deterministically, meanwhile it also simplifies some widely-adopted design choices in conventional NeRF rendering.

\textbf{Skipping Empty Spaces.} In NeRF rendering, we are supposed to focus only on the {\em surface} of scene objects. As shown in Fig.~\ref{fig:kd-voxel}, all our K-D voxels are produced to stick to the surface of objects, which the point cloud is constructed to describe. Such a property allows us to avoid explicitly ``skipping'' empty spaces, which often requires extra consideration in most NeRF models -- only considering the space inside a voxel automatically focuses on the surface; as the depth of the voxel's node goes deeper, it becomes closer to the surface. Moreover, during the construction of the \mbox{K-D} tree, the points at each node are divided within its sub-nodes, and the node's voxel fully covers all its sub-nodes. This further provides us with a native {\em top-down recursive} procedure to locate the voxels intersected with the querying ray: We start from the root node, and recurse on the sub-nodes until we (1) reach a pre-set node depth (or equivalently, a pre-set voxel size) and then query within the associated voxel, or (2) stop recursion on non-intersected nodes.

\begin{figure*}[t!]
\centering
\vspace{-3mm}
\centerline{\includegraphics[width=.7\linewidth]{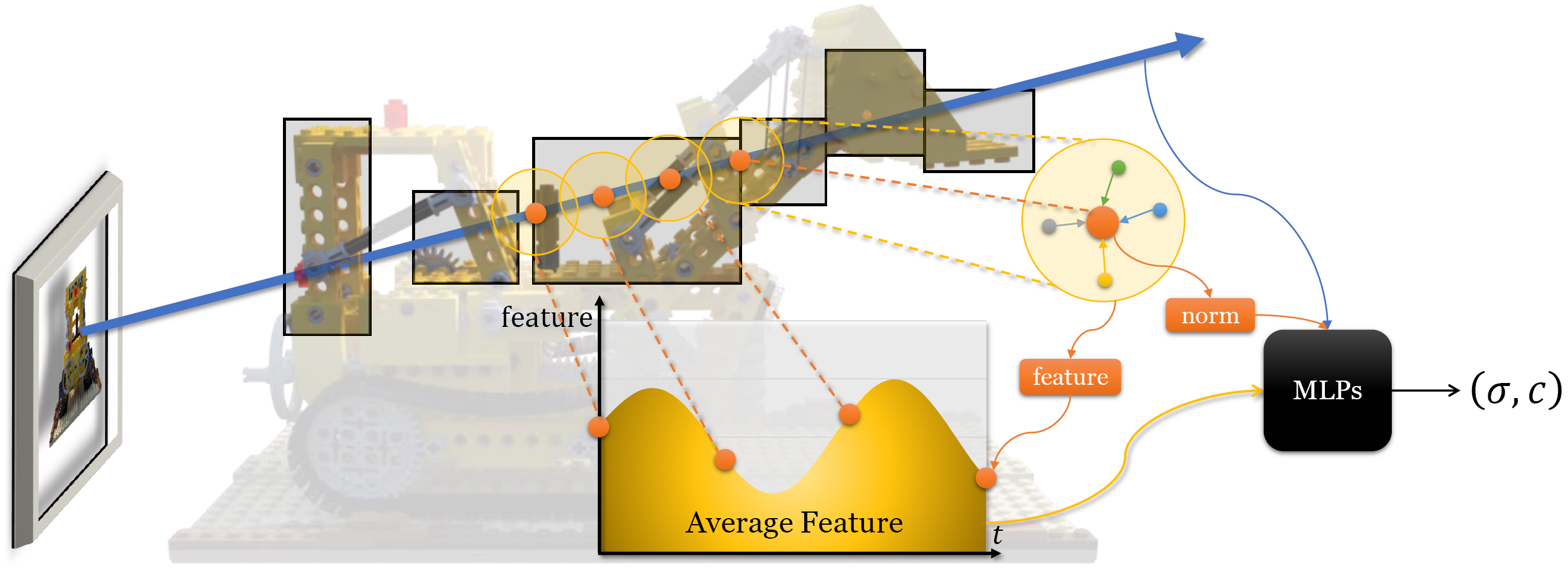}}
\vspace{-3mm}
\caption{\textbf{Our \themodel architecture.} We propose deterministic spline integration for KNN-based point features over each K-D tree-guided density-adaptive voxel, and model the color via Phong reflection with normal vectors estimated from the point cloud's shape.}
\vspace{-6mm}
\label{fig:render}
\end{figure*}

\textbf{Density-Adaptive Rendering.} An important design in NeRFs is the coarse-to-fine strategy for density-adaptive rendering, so that more points are sampled in high-volume density areas. Our K-D voxels natively support such a design {\em without additional bells and whistles}. This is because voxels in the same K-D tree layer contain the same number of points. As the point density can be regarded as an approximation of volume density, all such voxels have the same density. Therefore, we directly use K-D voxels to guide the density-adaptive rendering. Specifically, we conduct the rendering process at the voxels of some bottom layers in the K-D tree. For each querying ray, we use the aforementioned recursive procedure to locate the minimal intersected voxels that are deep enough. Here ``minimal'' means that the ray intersects with the node's voxel, but does not intersect with any sub-node's voxel. These intersected voxels divide the querying ray into several segments (Fig.~\ref{fig:render}). The segments covered by a voxel are close to the surface and used for rendering, while those not covered are in empty spaces.

\textbf{Deterministic Spline Integration.}
DIVeR~\cite{diver} shows that deterministic integration outperforms stochastic integration in NeRF rendering. So
we perform a deterministic integration to obtain the segment's feature within each voxel. Since we do not necessarily have points at the voxel's vertices, the trilinear interpolation used in DIVeR is not feasible here. Instead, we use spline integration. For the $i$-th intersected voxel in the ray passing order, we uniformly select points in this segment, and integrate their features to obtain the average feature $f_i$ of the segment of the $i$-th voxel:
\vspace{-1.2mm}
\begin{equation}
    f_i = \frac{1}{r_i-l_i} \int_{l_i}^{r_i} \textbf{feature}(o+t\cdot d) \mathrm{d}t,
\end{equation}\vspace{-3mm}

\noindent where $[l_i,r_i]$ is the intersection interval, and $o$ and $d$ are the source and direction of  the querying ray, respectively. This average feature $f_i$ can be interpreted as the feature of {\em a representative point} $p_i$ located somewhere in the segment.

\textbf{KNN-Based Feature Aggregation.}
For each uniformly selected point $q$ in the segment during spline integration, we obtain its feature via weighted interpolation from the features of its $K$ nearest neighbors (KNN) in the point cloud:
\vspace{-1.2mm}
\begin{equation}\begin{aligned}
    &\textbf{feature}(q) = \sum_{p_j \in \mathrm{KNN}(q; K)} k_j e_j, \\
    &\{k_j\} = \mathop{\mathrm{SoftMax}}\limits_{p_j \in \mathrm{KNN}(q; K)} \left(\log \gamma_j - \log \|q-p_j\|_2^2\right),
    \label{eq:pts-fea-interpolate}
\end{aligned}\end{equation}\vspace{-3mm}

\noindent where for each point $p_j$ in the point cloud, we parameterize its confidence $\gamma_j$ and point feature $e_j$ as in PointNeRF.

\textbf{Phong Reflection-Based Color Modeling with Point Cloud Normal Vectors.} { To obtain the volume density $\sigma_i$ and color $c_i$ of the representative point $p_i$ in the $i$-th voxel,} we use the Phong reflection model~\cite{phong}. As we have the underlying point cloud, we use Open3D~\cite{open3d} to estimate the normal vector for each point, and integrate these vectors over the interval to get an average normal vector $n_i$. Such information better characterizes the shape of point clouds (scenes), which plays an important role in Phong-based rendering and also implicitly facilitates optimizing the point clouds (Sec.~\ref{sec:opt}). Consistent with RefNeRF~\cite{refnerf}, {we use multiple multilayer perceptrons (MLPs) to model other attributes, including volume density, tint, roughness, and diffuse and specular color, and use the Phong formula to calculate the final $c_i$ with these attributes.} Finally, we aggregate $c_i$ of all segments on the ray to obtain the final color:
\vspace{-1mm}
\begin{equation}
\begin{aligned}
    & c_\mathrm{pixel} = \sum_{i\ge 1} \tau_i \cdot (1 - \exp(-(r_i-l_i)\sigma_i) \cdot c_i,\\
    & \tau_i = \exp\left( -\sum_{i'=1}^{i-1} (r_{i'}-l_{i'}) \sigma_{i'} \right).
\end{aligned}
\end{equation}
\vspace{-4mm}

\subsection{Point Cloud Optimization}
\label{sec:opt}
\begin{figure}[t!]

\centering
\centerline{\includegraphics[width=.9\columnwidth]{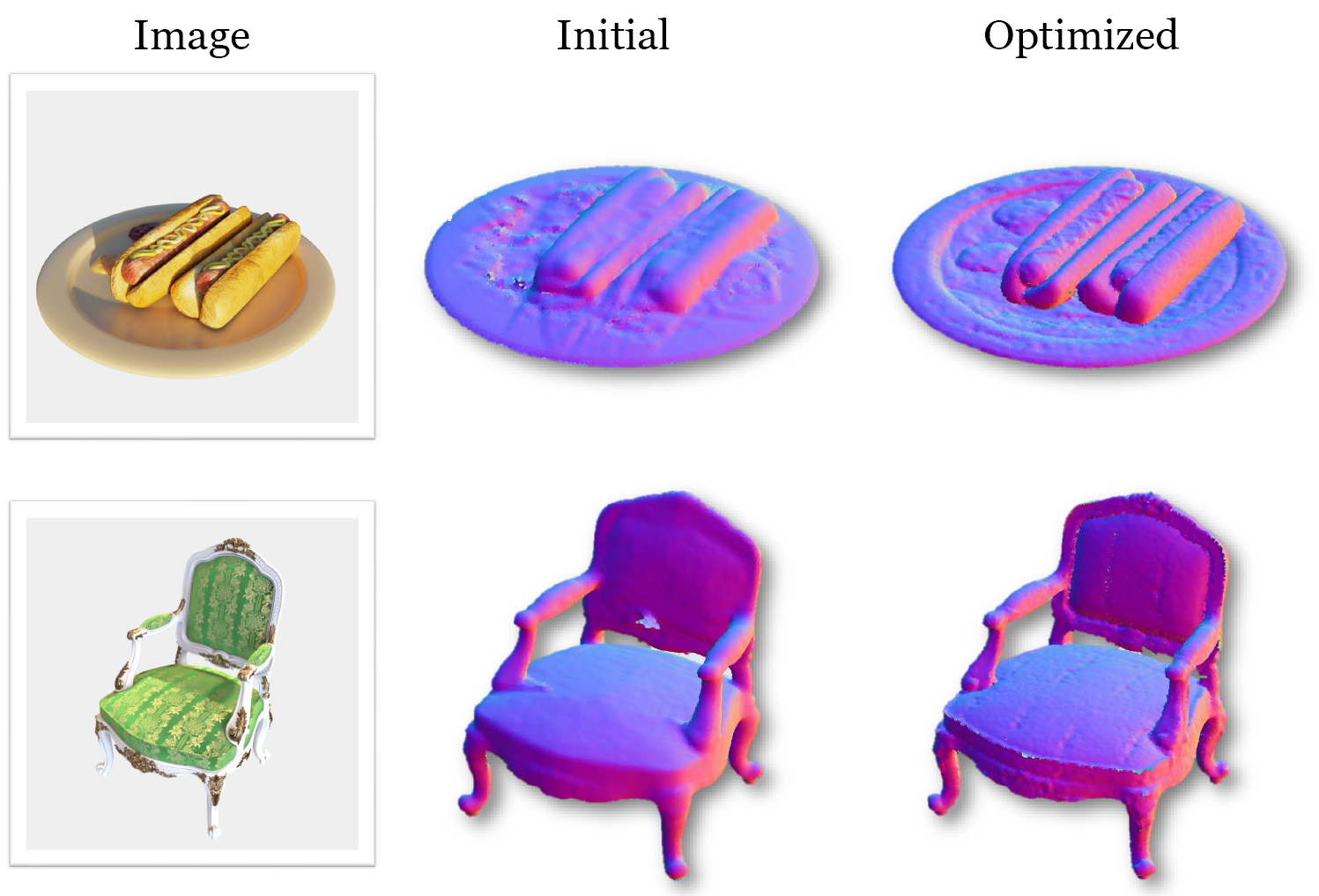}}
\vspace{-4mm}
\caption{In the two scenes of NeRF Synthetic~\cite{nerf}, \themodel optimizes the rough initial point cloud to a precise point cloud. The points are colored with their normal vectors.}
\label{fig:pg-init-final}
\vspace{-5mm}
\end{figure}

\textbf{Point Cloud Initialization.} To start training, we need a coarse initial point cloud. Consistent with PointNeRF, we use a point generation network, which consists of a multi-view stereo (MVS) model~\cite{deepmvs} based on a 3D convolutional neural network (CNN), to generate the points' coordinates and confidence values, and another 2D CNN~\cite{mvsnet} to generate their initial features. This network was pre-trained on the DTU training dataset~\cite{dtu}, and can generalize to other datasets and scenes. As shown in Fig.~\ref{fig:pg-init-final}, the initial point cloud generated by such a network is coarse and noisy. 

\textbf{Explicit Optimization via Pruning and Growing.} We perform a similar pruning and growing procedure as in PointNeRF, to prune outliers with low confidence $\gamma_j$ and fill holes in the point clouds. We make several important modifications over PointNeRF, and integrate this procedure with our deterministic integration (details in the supplementary).

\textbf{Implicit Optimization with Normal Vectors.} In addition to the explicit optimization, the point cloud is also optimized implicitly during training through the adjustment of point confidence $\gamma_j$. When computing the average normal vector for rendering, we aggregate normal vectors of nearby points weighted with their distance and confidence, where the confidence of noisy or inaccurate points with potentially abnormal normal vectors is adjusted accordingly. Moreover, we apply the normal vector regularization losses from RefNeRF~\cite{refnerf} to supervise the points' confidence w.r.t. their normal vectors.
These strategies collectively provide {\em implicit but more tailored} ways to optimize the point clouds. 
With both explicit and implicit optimization, \themodel obtains very precise point clouds (Fig.~\ref{fig:pg-init-final}).

%% file: 3.2-edit.tex
\section{Shape Editing with \themodel}
\label{sec:method-edit}
\label{sec:method-edit-v2}

\textbf{Formulation of General Shape Editing Tasks.}
We define the shape editing tasks based on {\em indexed} point clouds. To this end, we first re-define an indexed point cloud $P$ as a mapping from a point index $j$ to the corresponding point $p_j$,
\vspace{-1mm}
\begin{equation}
    P: j \to p_j,\!\text{~where~} p_j\in\mathbb{R}^3, j=1,\cdots,|P|.
\end{equation}
A shape editing task is defined as {\em another} indexed point cloud $Q(P)$,
\vspace{-1mm}
\begin{equation}
    Q(P): j \to q_j,\!\text{~where~} q_j\in\mathbb{R}^3\!\cup\!\{\varnothing\},j=1,\cdots,|P|,
\end{equation}
describing a shape editing task whether the $j$-th point moves from $p_j$ to $q_j$ or is deleted in the deformation if $q_j=\varnothing$. With this definition, $Q(P)$ can be an {\em arbitrary} point cloud with points properly matched to points in $P$ by same indices, regardless of connectivity or continuity. 

Our formulation represents a broad range of shape editing tasks. The mesh editing tasks in NeRF-Editing~\cite{nerfediting}, Deforming-NeRF~\cite{deformingnerf}, and CageNeRF~\cite{cagenerf} can be more simply and clearly defined here. For example, in NeRF-Editing, a mesh is exported from a general NeRF model, deformed manually, and converted to an ``offset'' or a continuous space deformation. We can depict such a task without ``offsets,'' by recording only the {\em final location} for {\em each point} without extra information. Notably, our formulation even models those whose deformation is {\em not continuous} in the space, {\eg, cutting a scene into two parts,} and thus cannot be covered and solved by NeRF-Editing, Deforming-NeRF, or CageNeRF, {as shown in the supplementary.}

\textbf{Editing Shape by Moving Points.} We design our shape editing scheme with \themodel. This is achieved by interpreting NeRF rendering as ``plotting'' the sampled points over the viewing ray. If we render the scene by naively plotting the point cloud $P$, the shape editing task can be addressed by replacing each point's coordinate from $p_j$ to $q_j$. For \themodel, we similarly replace the underlying point cloud from $P$ to $Q(P)$, while maintaining the confidence values and features. This method is general and can also be applied to any point-based NeRF model like PointNeRF.

\begin{figure}[t!]
\begin{center}
\centerline{\includegraphics[width=1.0\columnwidth]{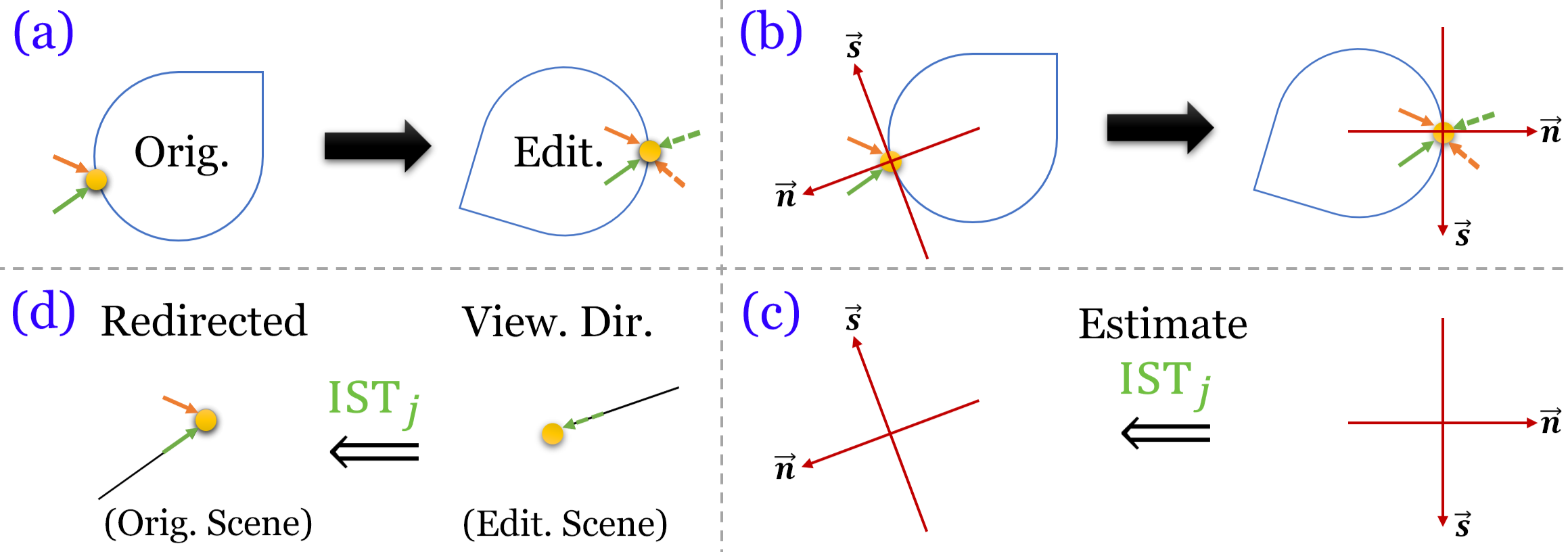}}
\vspace{-3mm}
\caption{\textbf{Infinitesimal surface transformation (IST).} \textcolor{blue}{(a)} As the view-dependent colors are modeled as absolute viewing directions, they (solid arrows at the right) are different from the correct colors (dashed arrows at the right) after deformation. We solve this by \textcolor{blue}{(b)} constructing a local coordinate system near the $j$-th point and \textcolor{blue}{(c)} modeling IST {\em from} the edited scene {\em to} the original scene with the coordinate systems, so as to \textcolor{blue}{(d)} redirect the viewing direction to the original scene when rendering the edited scene.}
\label{fig:viewdep-abs}
\vspace{-13mm}
\end{center}
\end{figure}

\textbf{Correcting View-Dependence with Infinitesimal Surface Transformation (IST).}
The editing method above can already obtain reasonable results. However, as illustrated in Fig.~\ref{fig:viewdep-abs}, the modeled view-dependent colors record the {\em absolute} viewing direction, making them incorrect after deformations that change their orientation.

To solve this issue, we model the infinitesimal surface transformation (IST) for each point to redirect the viewing ray in the correct direction. We construct a local coordinate system for each point to represent the orientation of the infinitesimal surface, using its normal vector and two point indices that are neighbors of the $j$-th point in both $P$ and $Q(P)$. By comparing these two coordinate systems, we can obtain an affine transformation $\mathrm{IST}_j$ for the $j$-th point to redirect the querying view direction (Fig.~\ref{fig:viewdep-abs}). This procedure is different from modeling space deformation~\cite{cagenerf,deformingnerf,nerfediting}, as we only need to model a simple affine transformation at each point, while those methods model a complicated, continuous, and non-linear deformation in the whole space.

Our proposed method requires a precise point cloud with normal vector-based color modeling. It is thus incompatible with PointNeRF, as PointNeRF is unable to obtain a desired point cloud to estimate the surface normal vectors. 

\textbf{Fine-Tuning on Deformed Scene.} 
Using the shape editing scheme introduced above, we can apply shape deformation on the scene modeled by our \themodel{} {\em without any modification} to the model architecture or rendering scheme, which means that the resulting model is still a valid, {fully functional} \themodel. Therefore, we can further fine-tune \themodel on the deformed scene if the ground truth is available. We can even fine-tune the infinitesimal surface transformation with other parameters, to rapidly adjust toward better ambient consistency. This makes \themodel desirable {\em in practice}, since in most applications, the final goal is not a zero-shot inference, but to {\em fit} the deformed scene well with reduced cost. By supporting fine-tuning, our \themodel aligns well with and achieves this goal.

As another point-based NeRF model, PointNeRF supports fine-tuning but cannot leverage infinitesimal surface transformation fine-tuning to further optimize the performance. On the other hand, mesh-based NeRF editing models~\cite{nerfediting,deformingnerf,cagenerf} do not support fine-tuning well: With deforming the space instead of the scene, these models' rendering scheme has highly changed. In their rendering process, a ray may go through a long, irregular way to reach the scene's surface. As the modeled space deformation might not be precise, it could be hard to tune the irregular space well, and even hurt other parts of the trained NeRF model. Such issues occur especially for some spaces with non-uniform density, since most of their model components (\eg, positional encodings, voxels) are not designed to deal with non-uniform spaces. Among all these methods, only our \themodel has complete support for fine-tuning.

%% file: 4-exp.tex
\section{Experiment}
\label{sec:exp}

\begin{figure}[t!]
\begin{center}
\centerline{\includegraphics[width=.95\columnwidth]{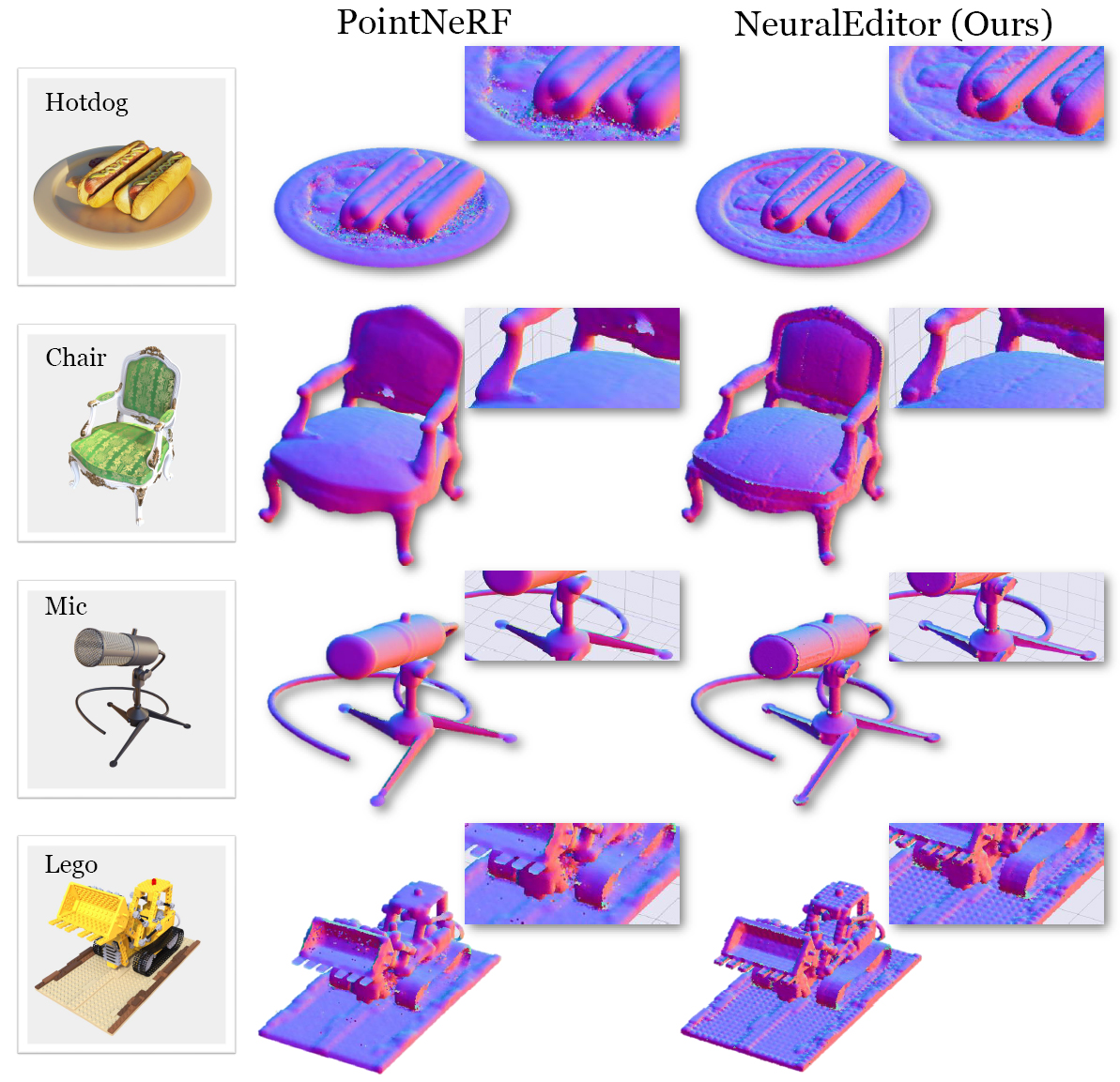}}
\vspace{-3mm}
\caption{\themodel generates much more precise point clouds than PointNeRF~\cite{pointnerf} in the four scenes of NeRF Synthetic~\cite{nerf}. The points are colored with their normal vectors. } 
\label{fig:exp-cloud}
\vspace{-9mm}
\end{center}
\end{figure}

\textbf{Point Cloud Generation.}
The underlying point cloud is fundamental to all editing tasks. Fig.~\ref{fig:exp-cloud} first provides a qualitative comparison of point clouds generated by our \themodel and PointNeRF~\cite{pointnerf} on NeRF Synthetic~\cite{nerf}. Ours are much more precise with sharper details, \eg, the mayonnaise on the Hotdog's sausage, the uneven texture on the Chair's cushion, the edge of the Mic's stand, and the Lego brick's studs. By contrast, PointNeRF's point clouds are blurred and noisy, lose most of the details, and even contain obvious shape deflects on the Hotdog's plate and Chair's backrest. This shows that while the point cloud generation task is challenging, \themodel generates a super-precise point cloud which is crucial for shape editing tasks.

\begin{figure}[t!]
\begin{center}
\centerline{\includegraphics[width=1\columnwidth]{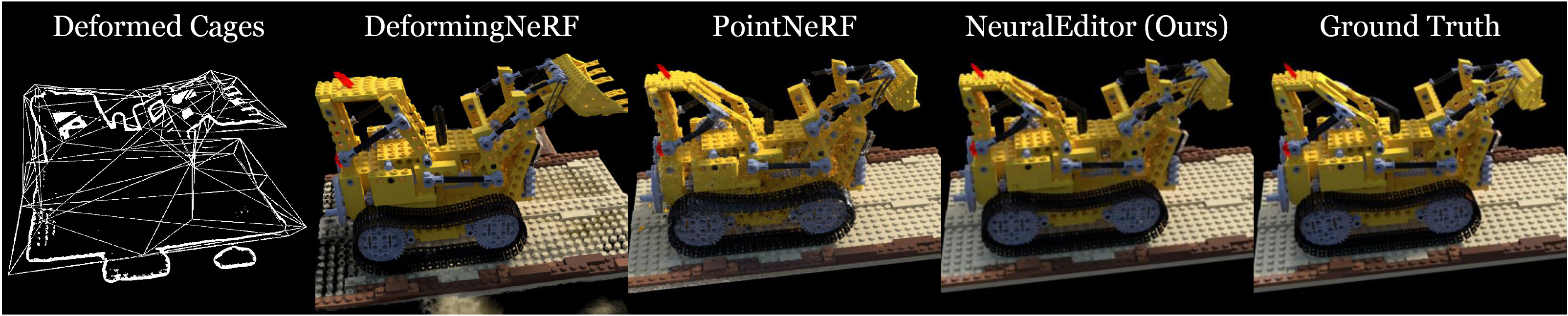}}
\vspace{-3mm}
\caption{With too coarse cages, DeformingNeRF~\cite{deformingnerf} is unable to perform the deformation faithfully, leading to poor results.}
\label{fig:exp-dnerf}
\vspace{-13mm}
\end{center}
\end{figure}

\begin{table*}[ht]
\centering

\scalebox{0.7}{\begin{tabular}{l|cccccccc|cccccccc}
 \hline\hline
 \multirow{2}{*}{Model} & \multicolumn{8}{c|}{Zero-Shot Inference, PSNR $\uparrow$} & \multicolumn{8}{c}{Fine-Tune for 1 Epoch, PSNR $\uparrow$} \\
 \cline{2-17}
& Chair & Hotdog & Lego & Drums & Ficus & Materials & Mic & Ship & Chair & Hotdog & Lego & Drums & Ficus & Materials & Mic & Ship  \\
 \hline
 DeformingNeRF~\cite{deformingnerf} & 18.84 & - & 13.10 &- &- &- &- &- &- &- &- &- &- &- &- &-  \\
 PointNeRF~\cite{pointnerf}  &22.21 &25.95 &24.56 &21.00 &24.24 &21.21 &26.77 &21.19 &30.11 &36.08 &31.45 &27.16 &31.48 &27.55 &34.34 &28.90 \\
 Naive Plotting  &24.91 &27.01 &25.64 &21.29 &26.22 &21.65 &27.63 &22.29 &32.01 &36.38 &31.72 &28.09 &33.21 &30.31 &35.15 &30.01 \\
 \hline
 \themodel w/o IST  &24.92 &27.02 &25.65 &21.29 &26.24 &21.64 &27.64 &22.28 &32.24 &36.69 &32.79 &28.30 &33.34 &30.40 &35.28 &30.08\\
 \themodel (Ours)  &\textbf{25.85} &\textbf{27.49} &\textbf{27.46} &\textbf{21.84} &\textbf{27.19} &\textbf{23.18} &\textbf{27.75} &\textbf{24.16} &\textbf{32.53} &\textbf{37.22} &\textbf{32.95} &\textbf{28.35} &\textbf{33.53} &\textbf{30.82} &\textbf{35.46} &\textbf{30.44}\\

 \hline\hline
\end{tabular}}
\vspace{-3mm}
\caption{{ {\themodel} {\em significantly and consistently} outperforms PointNeRF and Naive Plotting on all deformed scenes of NeRF Synthetic in peak signal-to-noise ratio (PSNR), in both zero-shot inference and fine-tuning settings. Our infinitesimal surface transformation (IST) effectively improves the results by correcting the view-dependent colors. With the precise point cloud generated by \themodel, even Naive Plotting consistently outperforms PointNeRF. Comparison results under other metrics are in the supplementary.} }
\vspace{-4mm}

\label{tab:exp-deform-quan}
\end{table*}

\begin{figure*}[t!]
\begin{center}

\centerline{\includegraphics[width=1\linewidth]{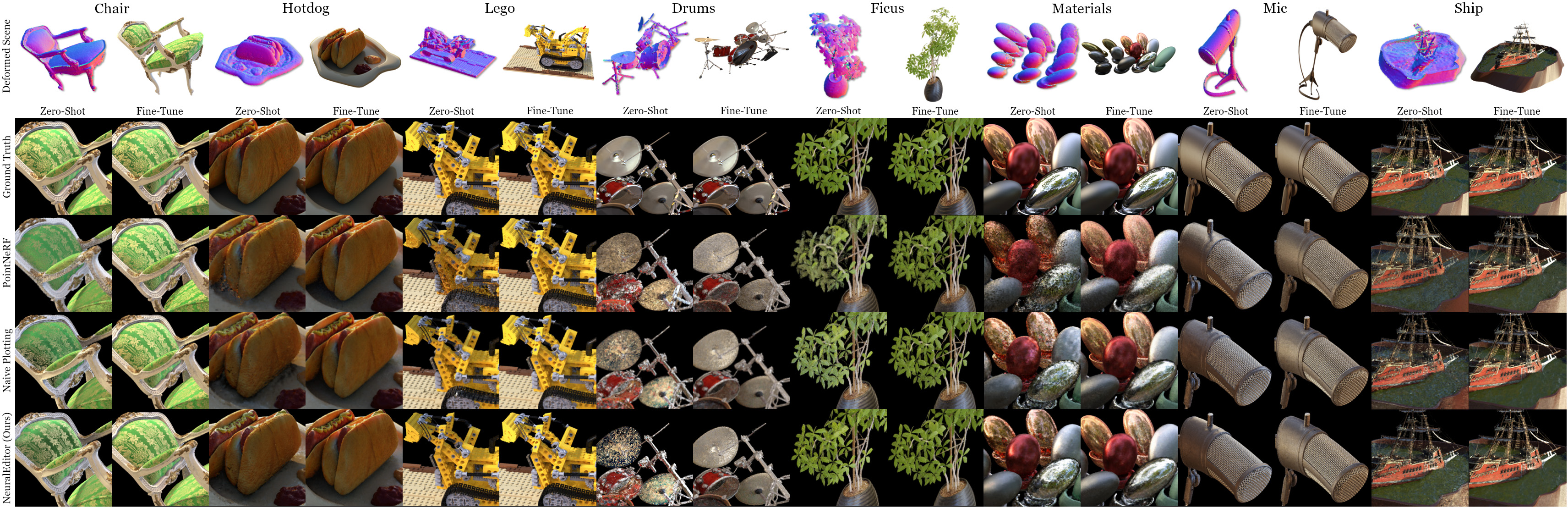}}
\vspace{-3mm}
\caption{{\themodel produces superior rendering results to PointNeRF, with significantly fewer artifacts in zero-shot inference. Fine-tuning further improves the {\em consistency of rendering with the ambient environment}. We use a black background for better visualization.} }
\label{fig:exp-all}
\vspace{-13mm}
\end{center}
\end{figure*}

\textbf{Experimental Settings.}
We mainly conduct experiments based on scenes from the NeRF Synthetic (NS) dataset. {NS is a widely-used NeRF benchmark constructed from Blender~\cite{blender} scenes. Due to the lack of publicly available benchmarks for shape editing, we use Blender to construct a reproducible benchmark, including the ground truth of edited scenes for evaluation and fine-tuning.} Our shape editing tasks cover {\em all eight scenes} in NS, while prior work~\cite{cagenerf,deformingnerf,nerfediting} only picks a few scenes. The provided images for NS scenes are with opacity, and there is no requirement for the background color. We evaluate and visualize the results on a black background, for better contrast and clearer detail visualization. {In the supplementary, we show the results on a white background with same conclusions.}

\textbf{Shape Editing Tasks.}
We evaluate our model on two types of shape editing tasks, as shown in Fig.~\ref{fig:teaser}:

\textit{(I) Shape (Mesh) Deformation Task.} {We consider the shape deformation task as in~\cite{nerfediting,cagenerf,deformingnerf}: deform the shape of a scene in a human-guided way. To construct our deformation tasks from NS and obtain the ground truth, we apply the shape deformation simultaneously to the scene and our point cloud within the provided Blender file. We perform both zero-shot inference and fine-tuning, and compare our rendering results with the ground truth. As domenstrated in Figs.~\ref{fig:exp-dnerf} and~\ref{fig:exp-all}, our deformation tasks are much more precise and {\em aggressive}, compared with those in previous work~\cite{nerfediting,cagenerf,deformingnerf}. In the supplementary, we also design a non-continuous deformation task and deformation tasks on the real-world dataset Tanks and Temples~\cite{tank} (with zero-shot inference only, as there is no ground truth available).}

\textit{(II) Scene Morphing Task.} We address a more challenging shape editing task that has {\em not} been investigated in prior NeRF editing work: the scene morphing task. Given two scenes $A$ and $B$, we should construct a path to gradually change one scene to the other, and the intermediate scenes should have reasonable appearances. We are required to render all intermediate scenes. For this task, we use the point cloud diffusion model~\cite{clouddiffusion} to generate intermediate point clouds with {\em latent space interpolation} between $A$ and $B$ as in \cite{clouddiffusion}, and we introduce a K-D tree-based~\cite{kdtree} matching algorithm to match the adjacent points to fix the indices of the intermediate scenes. To render an intermediate scene, we apply the shape transformation to the \themodel models trained for scenes $A$ and $B$, and then interpolate the rendering features to obtain the features of the intermediate scene for rendering.

\textbf{\themodel Variants.}
(1) Full \themodel: Our complete model with all components.
(2) \themodel~{\em without} infinitesimal surface transformation (IST): We remove the maintenance and optimization of infinitesimal surface transformation in scene editing. This key variant of \themodel enables us to evaluate the importance of IST as well as other components of our model. The {\em full ablation study} of \themodel is in the supplementary.

\textbf{Baselines.}
We compare \themodel against different types of baselines as follows.
(1) Naive Plotting: We use the point cloud generated by \themodel, computing each point's opacity and view-dependent colors with their point features. We render the scene by directly plotting/projecting the point cloud to the camera plane.
(2) PointNeRF~\cite{pointnerf}: For the shape deformation task, we apply the same deformation to the point clouds generated by PointNeRF. For the scene morphing task, we apply the matching algorithm to the point clouds generated by PointNeRF and the same intermediate point clouds generated by the point cloud diffusion model~\cite{clouddiffusion} for fairness.
(3) DeformingNeRF~\cite{deformingnerf}: DeformingNeRF is not compatible with the scene morphing task. For the shape deformation task, we perform the same deformation on vertices of given cages. Note that DeformingNeRF only released trained models for Lego and Chair, so we can only evaluate it on these two scenes.
While other models~\cite{cagenerf,nerfediting} support NeRF-based shape deformation via cages or exported meshes, we were unable to use them as baselines -- they did not provide executable code nor their deformed scenes for us to evaluate on their tasks.

\begin{figure*}[t!]
\centering

\centerline{\includegraphics[width=0.82\linewidth]{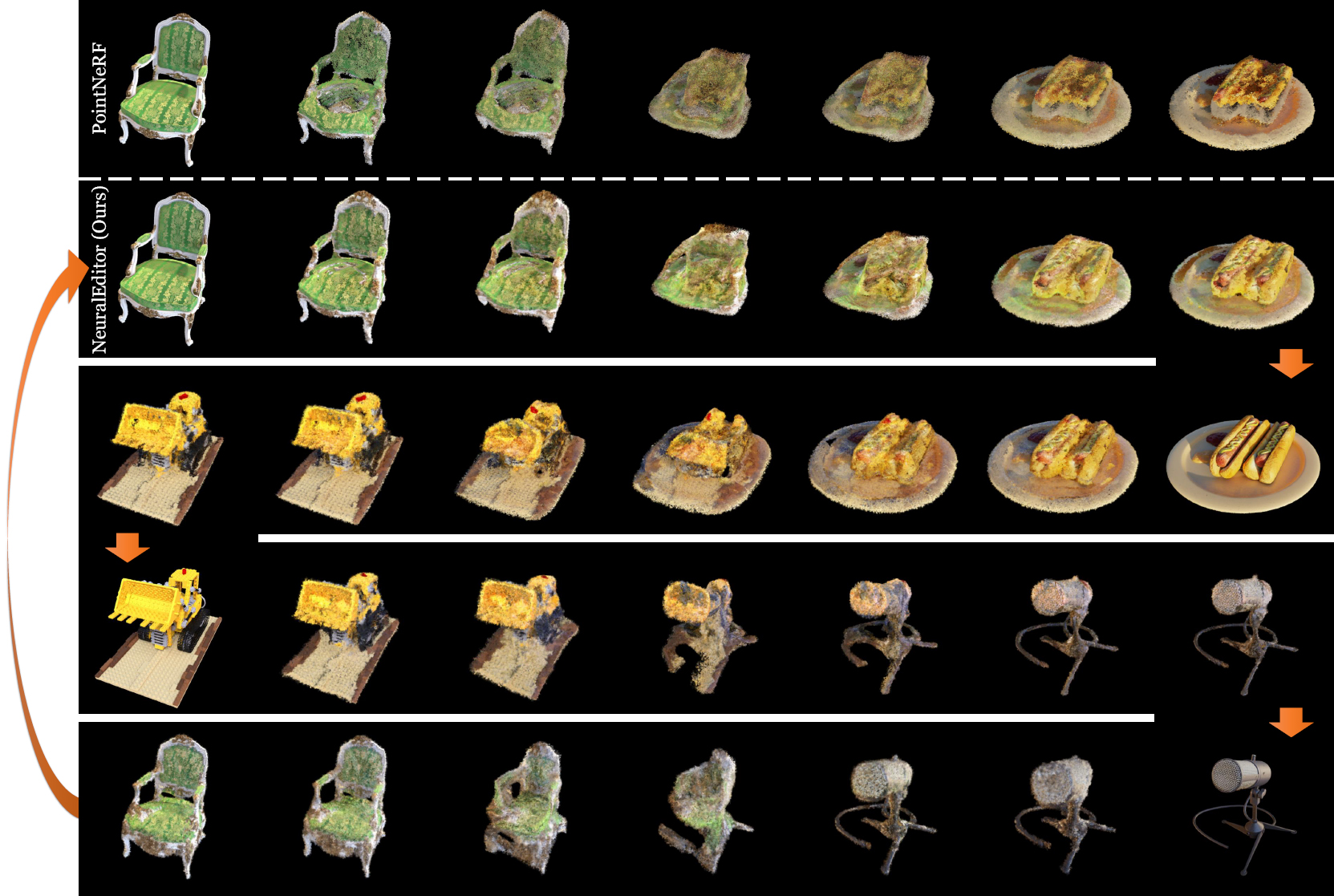}}
\vspace{-3mm}
\caption{Our \themodel produces {\em smooth morphing} results between Chair, Hotdog, Lego, and Mic in the NeRF Synthetic dataset, while PointNeRF produces results with blurry textures, black shadows, and gloomy, non-smooth colors. The rendering results in the looped morphing process are arranged in the shape of the numerical digit ``3,'' indicated by the dividing lines and arrows.}
\label{fig:exp-morph}
\vspace{-6mm}
\end{figure*}

\textbf{Shape (Mesh) Deformation Results.}{
The qualitative comparison is shown in Fig. \ref{fig:exp-all}. Both PointNeRF and Naive Plotting have many artifacts, like blur, wrong color, black or white shadows, noise, \etc, whereas our powerful \themodel produces clean and realistic rendering results. After fine-tuning, \themodel shows a significant improvement with better rendering results than PointNeRF, indicating that \themodel is able to achieve higher consistency with the surrounding ambient environment. Notably, in the Materials scene (the 6th scene from left), only our \themodel generates reasonable reflection, while both baselines show blurry and visually messy results. Also in the Drums scene, the bottom face of the gong is not visible in any of the training views in the original scene, so all models render poor results in zero-shot inference. However, after only fast fine-tuning, \themodel is able to precisely model the previously unknown surface and generate a substantially better result than PointNeRF, highlighting \themodel's strength in fast-fitting. All these results demonstrate that \themodel can handle various visual effects and make them consistent in the deformed scene. We provide the figure with higher resolution in the supplementary.

The quantitative comparison is summarized in Table \ref{tab:exp-deform-quan}. We observe that: 
(1) Our \themodel consistently outperforms all the baselines and variants for both zero-shot inference and fine-tuning settings. 
(2) With the precise point cloud generated by \themodel, the Naive Plotting baseline even consistently outperforms PointNeRF. 
(3) Our `w/o IST' variant has a comparable performance to Naive Plotting with the same point cloud and features in the zero-shot inference setting, but after fine-tuning its performance is significantly higher than Naive Plotting, validating the capability of \themodel in NeRF modeling. 
}

Notably, DeformingNeRF~\cite{deformingnerf} performs poorly in our benchmark with significantly lower metric values. As shown in Fig.~\ref{fig:exp-dnerf}, the cages provided by DeformingNeRF are too coarse, and cannot even cover the whole scene. Therefore, DeformingNeRF cannot faithfully perform the precise deformation in our benchmark, leading to poor rendering results. On the contrary, both PointNeRF and our \themodel at least faithfully perform the deformation, showing that point cloud is necessary for precise shape editing.

\textbf{Scene Morphing Results.}
The morphing results between 4 NeRF Synthetic scenes are shown in Fig.~\ref{fig:exp-morph}. The morphing process starts from Chair, morphs to Hotdog, Lego, Mic, and at last turns back to Chair. \themodel produces smooth rendering results on the point cloud diffusion-guided~\cite{clouddiffusion} intermediate scenes, mixing the textures of the two scenes in a reasonable way. In comparison, the rendering results produced by PointNeRF are unsatisfactory, with blurry textures, black shadows, and gloomy, non-smooth colors. These results show that our \themodel can render challenging intermediate morphing scenes and achieve decent results {\em with only the input of moved points}.

%% file: 5-conclusion.tex
\vspace{-1mm}
\section{Conclusion}
\label{sec:conclusion}
\vspace{-1mm}
This paper proposes \themodel, a point cloud-guided NeRF model that supports general shape editing tasks by manipulating the underlying point clouds.
Empirical evaluation shows \themodel to produce rendering results of much higher quality than baselines in a zero-shot inference manner, further significantly improving after fast fine-tuning. \themodel even supports smooth scene morphing between multiple scenes, which is difficult for prior work. We hope that our work can inspire more research on point cloud-guided NeRFs and 3D shape and scene editing tasks.

\noindent{\footnotesize{\textbf{Acknowledgement.} This work was supported in part by NSF Grant 2106825, NIFA Award 2020-67021-32799, the Jump ARCHES endowment, the NCSA Fellows program, the IBM-Illinois Discovery Accelerator Institute, the Illinois-Insper Partnership, and the Amazon Research Award. This work used NVIDIA GPUs at NCSA Delta through allocation CIS220014 from the ACCESS program. We thank the authors of NeRF~\cite{nerf} for their help in processing Blender files of the NS dataset. }

\normalsize

%% file: 6-appendix.tex
\twocolumn[{
\part*{\Large \begin{center}Supplementary Material\end{center}}

\vspace{-1em}

This document contains additional descriptions (\eg, implementation details, experimental setting details, \etc) and extra experiments (\eg, ablation study, deformation on the Tanks and Temples dataset, \etc). 
\vspace{1em}
}]
\appendix
\setcounter{table}{0}
\setcounter{figure}{0}
\renewcommand{\thefigure}{\Alph{figure}}
\renewcommand{\thetable}{\Alph{table}}
\renewcommand{\thealgorithm}{\Alph{algorithm}}

\section{Black/White Background}

In the main paper, we evaluated the models and presented the results on a black background, for better contrast and clearer detail visualization. Here, we provide experimental results for three representative deformed scenes of NeRF Synthetic~\cite{nerf} on a white background. As shown in Table~\ref{tab:exp-deform-quan-white}, we have the same conclusions as the experiment on a black background: {\themodel} {\em significantly and consistently} outperforms PointNeRF~\cite{pointnerf} and Naive Plotting, in both zero-shot inference and fine-tuning settings.

Interestingly, we observe that the results on a white background have consistently worse metric values than those on a black background. We find that this phenomenon is {\em not specific} to our task of shape deformation. In fact, even in rendering the original scene, we observe that a white background results in lower metric values, as shown in Table \ref{tab:bgcolor} for the experiment of PointNeRF 20K (PointNeRF trained for 20K epochs, following an evaluation setting in \cite{pointnerf}) on Lego of NeRF Synthetic. The impact of the background color on NeRF rendering is an interesting aspect for future investigation.

\begin{table}[t!]
\centering

\scalebox{0.7}{\begin{tabular}{l|ccc|ccc}
 \hline\hline
 \multirow{2}{*}{Model} & \multicolumn{3}{c|}{Zero-Shot Inference} & \multicolumn{3}{c}{Fine-Tune for 1 Epoch} \\
 \cline{2-7}
& Hotdog & Lego & Drums & Hotdog &Lego &Drums \\
 \hline
\multicolumn{1}{c|}{} & \multicolumn{6}{c}{PSNR $\uparrow$}\\
 \hline
 
 DeformingNeRF~\cite{deformingnerf} & - & 12.28 &- &- &- &- \\
 PointNeRF~\cite{pointnerf} & 25.48& 23.84& 20.52& 35.71& 30.10& 26.71 \\
 Naive Plotting & 26.87& 24.70& 20.92& 36.00& 30.94& 27.45\\
 \themodel w/o IST & 26.88& 24.71& 20.92& 36.27& 31.45& 27.64\\
 \themodel (Ours) & \textbf{27.27}& \textbf{26.13}& \textbf{21.41}& \textbf{36.66}& \textbf{31.70}& \textbf{27.68}\\
 \hline
\multicolumn{1}{c|}{} & \multicolumn{6}{c}{SSIM $\uparrow$}\\
 \hline
  DeformingNeRF & - & 0.690 &- &- &- &- \\
 PointNeRF & 0.948& 0.932& 0.902& 0.987& 0.976& 0.955\\
 Naive Plotting & 0.951& 0.932& 0.913& 0.987& 0.979& 0.963\\
 \themodel w/o IST & 0.951& 0.932& 0.913& 0.987& 0.981& 0.963\\
 \themodel (Ours) & \textbf{0.957}& \textbf{0.964}& \textbf{0.920}& \textbf{0.988}& \textbf{0.983}& \textbf{0.964}\\
  \hline
\multicolumn{1}{c|}{} & \multicolumn{6}{c}{LPIPS AlexNet $\downarrow$}\\
 \hline
  DeformingNeRF & - & 0.271 &- &- &- &- \\
 PointNeRF & 0.072& 0.064& 0.107& 0.033& 0.025& 0.067\\
 Naive Plotting & 0.066& 0.055& 0.086& 0.022& 0.019& 0.044\\
 \themodel w/o IST & 0.064& 0.054& 0.085& \textbf{0.021}& 0.017& \textbf{0.043} \\
 \themodel (Ours) & \textbf{0.059}& \textbf{0.031}& \textbf{0.079}& \textbf{0.021}& \textbf{0.016}& \textbf{0.043}\\
  \hline
\multicolumn{1}{c|}{} & \multicolumn{6}{c}{LPIPS VGG $\downarrow$}\\
 \hline
  DeformingNeRF & - & 0.291 &- &- &- &- \\
 PointNeRF &  0.079& 0.088& 0.108& 0.053& 0.054& 0.080\\
 Naive Plotting & 0.080& 0.089& 0.093& 0.047& 0.049& 0.062\\
 \themodel w/o IST & 0.079& 0.087& 0.092& 0.045& 0.043& 0.061\\
 \themodel (Ours) & \textbf{0.073}& \textbf{0.056}& \textbf{0.087}& \textbf{0.044}& \textbf{0.041}& \textbf{0.060}\\
 \hline\hline
\end{tabular}}
\caption{Consistent with the results on a {\em black} background in the main paper, we have the same conclusions when using on a {\em white} background here: {\themodel} {\em significantly and consistently} outperforms PointNeRF~\cite{pointnerf} and Naive Plotting on the three {representative} deformed scenes of NeRF Synthetic~\cite{nerf}, in both zero-shot inference and fine-tuning settings. With the precise point cloud generated by \themodel, even Naive Plotting consistently outperforms PointNeRF. The metrics investigated here are peak signal-to-noise ratio (PSNR), structural similarity index measure (SSIM), and learned perceptual image patch similarity (LPIPS).}

\label{tab:exp-deform-quan-white}
\end{table}

\begin{table}[t!]
\centering
\scalebox{1}{\begin{tabular}{l|c}
 \hline
  Background Color & PSNR~$\uparrow$ \\
  \hline
  White~\cite{pointnerf} &32.40 \\
  Black &32.99 \\
  \hline
  
\hline
\end{tabular}}

\caption{For the conventional rendering~\cite{nerf} of the original scene of NeRF Synthetic (\eg, Lego here), the metric values are also slightly worse when using a white background as in prior work~\cite{nerf,pointnerf} than a black background.}
\label{tab:bgcolor}
\end{table}

\section{Additional Ablation Study}
\label{sec:ablation}

\begin{table*}[t!]
\centering
\scalebox{1}{\begin{tabular}{c|l|cc}
 \hline
  \multirow{2}{*}{Type} & \multirow{2}{*}{Variant} &\multicolumn{2}{c}{PSNR $\uparrow$ on Hotdog}  \\
  \cline{3-4}
 & & Zero-Shot & Fine-Tune \\
 \hline\hline
 \multirow{2}{*}{NeRF Model} & \textbf{Full \themodel}  & \textbf{27.49} & \textbf{37.22}  \\
  & $-$\,our improved point cloud-guided NeRF \ $+$\,PointNeRF~\cite{pointnerf} & 25.95 & 36.08\\
 \hline\hline
 \multirow{7}{*}{Component} & \textbf{Full \themodel}  & \textbf{27.49} & \textbf{37.22}  \\
 \cline{2-4}
 &  $-$\,IST (`Ours w/o IST') & 27.02 & 36.69 \\
 \cline{2-4}
 &  $-$\,integration \ $+$\,traditional point sampling & 27.48 & 36.69 \\
 \cline{2-4}
 &  $-$\,deterministic integration \ $+$\,stochastic integration & 27.46  & 36.87 \\
 \cline{2-4}
 &  $-$\,NeRF modeling \ $+$\,plotting (`Naive Plotting') & 27.01 & 36.38 \\
 \cline{2-4}
 &  $-$\,normal vectors & 27.26 & 37.00 \\
 \cline{2-4}
 &  $-$\,Phong reflection color modeling \ $+$\,traditional color modeling & 27.21 & 36.51 \\
 \hline\hline
 \multirow{4}{*}{Point Cloud} & \textbf{Full \themodel}  & \textbf{27.49} & \textbf{37.22}  \\
 \cline{2-4}
 & $-$\,point cloud optimization (w/ initial point cloud) & 25.56 & 35.93 \\
  \cline{2-4}
 & $-$\,our optimized point cloud \ $+$\,point cloud optimized by PointNeRF  & 26.61 & 35.68\\
  \cline{2-4}
 & $-$\,50\% points & 26.84 & 36.59 \\
 \hline\hline

 \multirow{5}{*}{Fine-Tune} & Fine-tune 0 epoch (`zero-shot')  & \multicolumn{2}{c}{27.49}  \\
 \cline{2-4}
 & Fine-tune 1 epoch & \multicolumn{2}{c}{37.22}  \\
 \cline{2-4}
 & Fine-tune 4 epochs & \multicolumn{2}{c}{38.12}  \\
 \cline{2-4}
 & Fine-tune 10 epochs & \multicolumn{2}{c}{38.56}  \\
 \cline{2-4}
 & Fine-tune 10 epochs w/ point cloud optimization & \multicolumn{2}{c}{38.87}  \\
 \hline\hline
\vspace{-1em}
\end{tabular}}

\caption{Ablation study experiments show that (1) All components in \themodel benefit the rendering results on deformed scenes; (2) Our \themodel generates precise point clouds, which are crucial for shape editing tasks; (3) Our \themodel even supports point cloud optimization during fine-tuning to further improve the rendering performance. `$-$' denotes that a certain component is removed, while `$+$' denotes that a certain component is added.}
\label{tab:exp-ablation}
\end{table*}

Our additional ablation study results are in Table \ref{tab:exp-ablation}. We observe that:

\begin{itemize}
    \item All components in our \themodel, including infinitesimal surface transformation (IST), deterministic integration, and Phong reflection, benefit the rendering results. Notably, our \themodel still outperforms PointNeRF {\em without} any of these components.
    \item For the variant without Phong reflection modeling, the zero-shot performance is close to that of the full \themodel, but the performance gap becomes much larger after fine-tuning. This demonstrates that the use of visual attributes modeled by Phong reflection (\eg, tint) helps fast fitting to the ambient environment.
    \item Our \themodel's performance drops with the initial point cloud or the final point cloud produced by PointNeRF, showing the importance of a precise point cloud optimized by our \themodel for the rendering task. In these experiments, we apply de-noising on the point clouds to support IST, which is not compatible with the noisy original point clouds generated by PointNeRF.
    
    \item With half of the points, the performance of \themodel decreases, but it still outperforms the baseline PointNeRF. This validates that it is not the model capacity but our model design together with the precise point clouds that leads to our superior performance.
    \item Our \themodel even supports point cloud optimization on the deformed scene, which further improves the rendering results and achieves better PSNR than fine-tuning without point cloud optimization for the same epochs. The detailed settings of fine-tuning are described in Section \ref{sec:training-settings}.
\end{itemize}

\section{Non-Continuous Deformation Task}

\begin{figure}[t!]
\begin{center}
\centerline{\includegraphics[width=\linewidth]{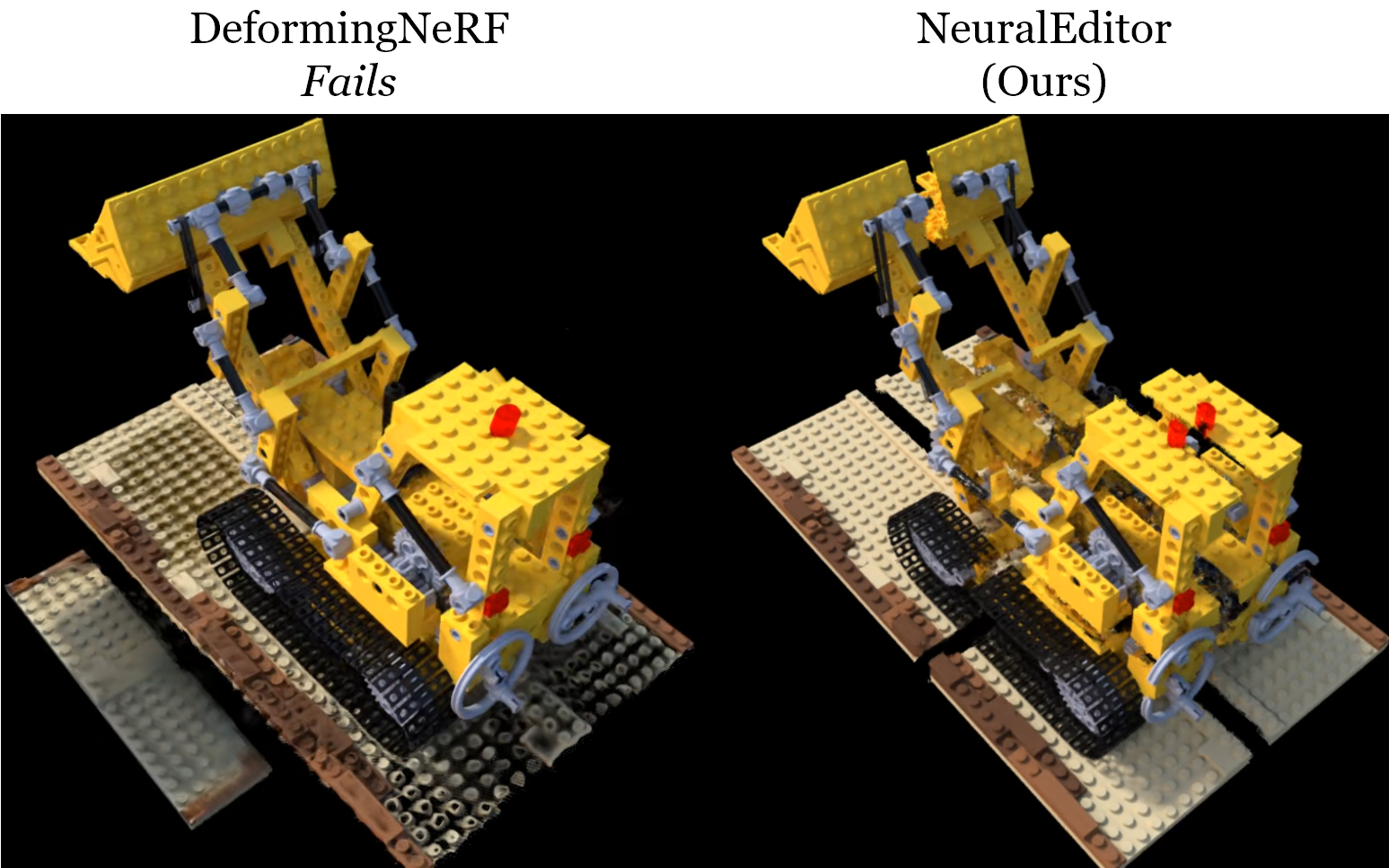}}
\caption{{Unlike Deforming-NeRF~\cite{deformingnerf}, our \themodel has native support for non-continuous deformation tasks.} } 
\label{fig:exp-noncont}
\vspace{-1em}
\end{center}
\end{figure}

{Figure \ref{fig:exp-noncont} shows a non-continuous deformation task constructed on the Lego scene of NeRF Synthetic, by cutting the scene from the $x\mathrm{O}y$, $y\mathrm{O}z$, and $z\mathrm{O}x$ planes. \themodel natively supports such deformation, while DeformingNeRF fails, further validating the superiority of \themodel over cage-based methods for tackling deformation tasks.}

\section{Experiment on Tanks and Temples Dataset}

\label{sec:tanks}

\begin{figure*}[t!]
\centering
\centerline{\includegraphics[width=1\linewidth]{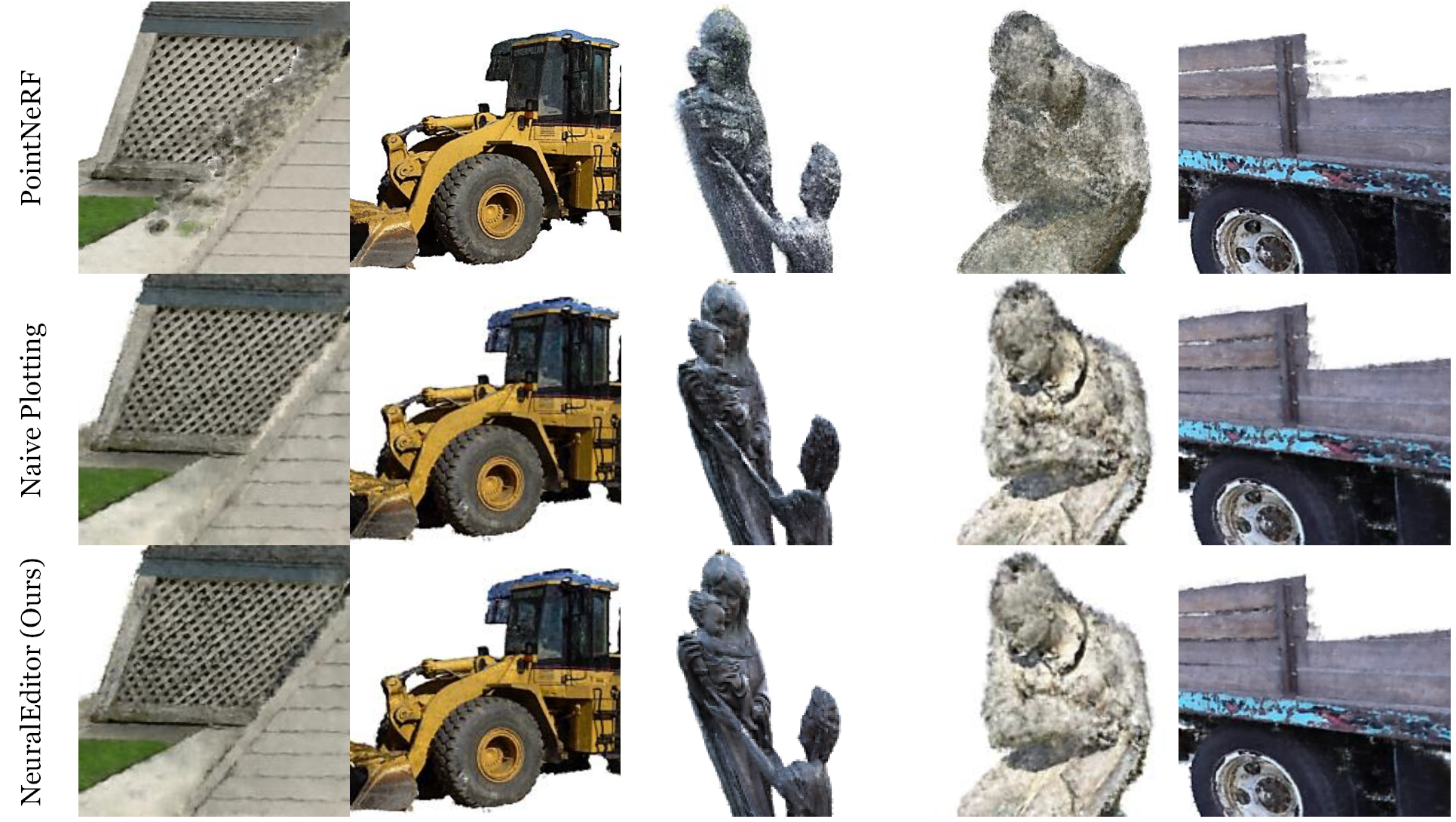}}
\caption{Our \themodel also generates better rendering results on deformed scenes of the Tanks and Temples~\cite{tank} dataset with much fewer artifacts.}
\label{fig:highres-exp-tank}
\vspace{-1em}
\end{figure*}

We also present the evaluation results of our \themodel and baselines Naive Plotting and PointNeRF~\cite{pointnerf} on the Tanks and Temples dataset~\cite{tank}, as shown in Figure~\ref{fig:highres-exp-tank}. As this dataset provides white-background ground truth images for original scenes, we use a white background for evaluation and visualization. Our \themodel still produces better rendering results with fewer artifacts than baselines.

Note that Tanks and Temples is not a standard NeRF dataset but a multiview stereo (MVS) dataset. It contains some background regions that are not fully cut out, \eg, the blue bar on the top of Caterpillar's cab (2nd column in Figure~\ref{fig:highres-exp-tank}) is a part of the sky background, making the NeRF rendering results blurry and noisy as some points are mistakenly grown in those regions. We propose a background sphere technique (explained in Section~\ref{sec:limit-pnerf}) to improve the rendering results in this situation. While this approach is helpful, it cannot completely resolve the issue. Additionally, inconsistent exposure settings used across different views of the same scene cause the rendering results to appear blurry and flashing. Therefore, all three methods cannot achieve rendering results as clean and realistic as those on NeRF Synthetic, but ours still significantly outperforms the baselines.

\section{Intermediate Point Clouds for Scene Morphing}

We show the intermediate point clouds for the scene morphing task (Figure 9) in Figure \ref{fig:morph-cloud}.

\section{Other Metrics in Table 1}

We provide the results of the shape deformation task (Table 1 in the main paper) under the metrics of peak signal-to-noise ratio (PSNR), structural similarity index measure (SSIM), and learned perceptual image patch similarity (LPIPS) in Table~\ref{tab:exp-deform-complete}.

\section{High-Resolution Visualization Figures}

Here we provide the higher-resolution versions of the same experimental visualization figures in the main paper. The correspondence is listed below: 

\begin{itemize}
    \item Figure 6 (optimized point clouds): Figure \ref{fig:highres-exp-cloud}.
    \item Figure 7 (baseline DeformingNeRF): Figure \ref{fig:highres-exp-dnerf}.
    \item Figure 8 (shape deformation): Figure \ref{fig:highres-exp-all}.
    \item Figure 9 (scene morphing): Figure \ref{fig:highres-exp-morph}, with full morphing results for the baseline PointNeRF.
\end{itemize}

\section{Implementation Details}

\subsection{K-D Voxels \& Integration}
\label{sec:kdvoxels}

We use a 21-layer K-D tree to build K-D voxels, which can contain up to $2^{21}$ ($\sim$ 2 million) points. Compared with PointNeRF that typically deals with 1 million points, our \themodel can hold twice the number of points for higher capacity, and also achieves better results with the same magnitude of points as PointNeRF, as shown in Section \ref{sec:ablation}.

We use the voxels in the bottom 4 non-leaf layers for rendering, and use the additional 5th layer from the bottom only for point cloud growing. When doing integration for rendering, we uniformly select $s$ points on the intersect interval, including the two side points, and apply spline integration using the features of these points, where we choose $s=8,11,16,22,22$ for the 5 used layers from the bottom. For the feature of each selected point, we use its $K$ nearest neighbors (KNN) for interpolation, where $K=32$. 

Such integration is not only for aggregating the average point features over the interval, but also for calculating the average normal vectors and average IST -- we use KNN to interpolate these values in a similar manner as interpolating the features. For each segment, we use the average point feature, average normal vector, and average IST in the color modeling of the representative point.

\subsection{Phong Reflection Color Modeling}
\label{sec:color-model}

Consistent with RefNeRF~\cite{refnerf}, we use several multilayer perceptrons (MLPs) that directly take the integrated average feature as input to obtain the tint, roughness, modeled normal vector, and diffuse and specular color. The modeled normal vectors are trained with and controlled by the regularization losses introduced in RefNeRF. These normal vectors may be different from the normal vectors estimated from the point cloud, and they are optimized towards the estimated normal vectors via an additional regularization loss weighted by point confidence. The effect of this somewhat redundant modeling of normal vectors is two-fold: (1) The outlier points with abnormal normal vectors can be difficult to fit into the modeled ones, so their confidence values will be driven to zero when minimizing the regularization loss; (2) The estimated normal vectors can supervise and provide extra shape information for the point features through the MLP that models the normal vectors.

\subsection{Point Cloud Optimization}
\label{sec:cloud-opt}
\textbf{Pruning \& Growing.} Following PointNeRF, we define the defects of a point cloud as two types: {\em outliers} and {\em holes}. So any shape defection condition can be decomposed into a sequence of these two types of simple defects. We design our optimization strategy based on the ``pruning and growing'' (P\&G) method introduced in PointNeRF, which prunes the outliers by driving their confidence values to zero during training, and grows the point cloud to probe holes by selecting points from the sampled points on the training rays. 

Our strategy is different from PointNeRF in several important ways. In our growing process, we grow all rays in the training dataset at a special evaluation epoch (``growing epoch''). For each ray, we pick the point far from any point in the point cloud with the highest volume density as a candidate. The candidate point whose ray has a higher pixel rendering loss is assigned a higher priority for being added to the point cloud. We then introduce a K-D tree-guided algorithm to down-sample these growing candidates: We start from the root node, and recurse on the sub-nodes until we reach a pre-set voxel size; we then add the candidate point with the highest priority in this voxel to the point cloud, while disregarding other candidate points in the same voxel. Doing so ensures the added candidate points uniformly distributed throughout the space, by preserving only the points with the highest priorities within their respective regions. We finalize the growth of candidate points by setting their features to interpolated features and their confidence values to $0.5$. Note that PointNeRF uses P\&G in conjunction with a stochastic training process, whereas we apply it to our deterministic \themodel during a standalone growing epoch, which enhances training stability. As a result, we obtain precise point clouds while PointNeRF cannot.

\textbf{Point Cloud Denoising.} It is common that a point cloud contains some noisy or isolated points. To remove these isolated points, we use Open3D~\cite{open3d} to identify statistical outliers w.r.t. the $K$-th nearest neighbor distance of each point, where $K=64$ in our setting. Also, as mentioned above, noisy points may have irregular estimated normal vectors, and their confidence will be driven to zero with the regularization losses introduced in Section \ref{sec:color-model}.

\textbf{Point Cloud Optimization Process.} We regard one ``point cloud optimization process'' as an additional process before one training epoch, which includes (1) applying one growing epoch to obtain grown points, (2) pruning the isolated points and the points with confidence values lower than $0.1$, and (3) constructing K-D voxels with the adjusted point cloud for further training. During training (Section \ref{sec:training-settings}), we apply this process with a few training epochs to optimize the point cloud.

\subsection{Training Settings}
\label{sec:training-settings}
\textbf{Training on Original Scene.} We pre-train \themodel on the original scene using per-scene optimization, and obtain a model that includes a precise point cloud and its points' features for shape editing tasks. During per-scene pre-training, we start with the initial point cloud generated by the point generation network. We first train our model for 3 epochs as warm-up. From the 4th to the 12th epoch, we include an additional point cloud optimization process (Section \ref{sec:cloud-opt}) before each training epoch. By the end of the 12th epoch, the point cloud is determined and precise enough. We keep tuning the model parameters on this underlying point cloud for up to 100 epochs, controlled with early stopping. Notably, the precise point cloud can be determined and obtained within the first 12 training epochs, even though the whole training process may take a long time.

\textbf{Fine-Tuning on Deformed Scene.} We apply the same training process during fine-tuning on deformed scenes. For the setting ``Fine-tune 10 epochs w/ point cloud optimization'' in Table~\ref{tab:exp-ablation}, we first train one epoch as warm-up, then train with an additional point cloud optimization process for 5 epochs, and continue to fine-tune on the optimized point cloud for the rest 4 epochs.

\subsection{Matching Algorithm for Scene Morphing}

{
To perform scene morphing, we generate the intermediate point clouds using a point cloud diffusion model~\cite{clouddiffusion}. However, these point clouds are {\em unindexed}. To render the scene with \themodel, we need to assign an index to each point. This index assignment can be solved by a matching algorithm. Specifically, given point clouds $P_0, P_1,\cdots,P_n$, we can match the points in each adjacent pair of point clouds $P_{i}$ and $P_{i+1}$, and then permute the points in $P_{i+1}$ according to the matching to align the indices.

We design a simple matching algorithm based on K-D trees. We simultaneously build two K-D trees, one for each point cloud. At each node, we select the same division axis for the two point clouds, according to their union point set. At each leaf node, we match a pair of single points from each point cloud. Our algorithm aims to match points in the two point clouds that have similar relative locations. In an ideal case where there are numerous intermediate point clouds and every adjacent pair of point clouds is sufficiently close, this algorithm will lead to highly accurate matching results.
}

\section{Limitations}

\subsection{Point Cloud-Guided NeRF} 
\label{sec:limit-pnerf}
Despite offering several advantages over traditional NeRFs, the current point cloud-guided NeRF models, including ours and PointNeRF, still have some limitations. First, the scene is modeled with an explicit representation (the point cloud), which is not robust when modeling surfaces with complicated visual effects, \eg, a semi-transparent blurry mirror. When the model fails to interpret such visual effects, our point cloud optimization might not generate correct points close to the surface, limiting the model's ability to improve results. Additionally, these models cannot well support strategies in \mbox{NeRFReN}~\cite{nerfren} for simultaneously modeling the real-world scene and the mirrored scene to better handle mirror reflections. This is because a point cloud-guided NeRF model relies on MVS-based initialization and point cloud optimization that cannot accommodate such a form of co-optimization, which may significantly change the shape of the mirrored scene during training.

\begin{figure}[t!]
\begin{center}
\centerline{\includegraphics[width=\linewidth]{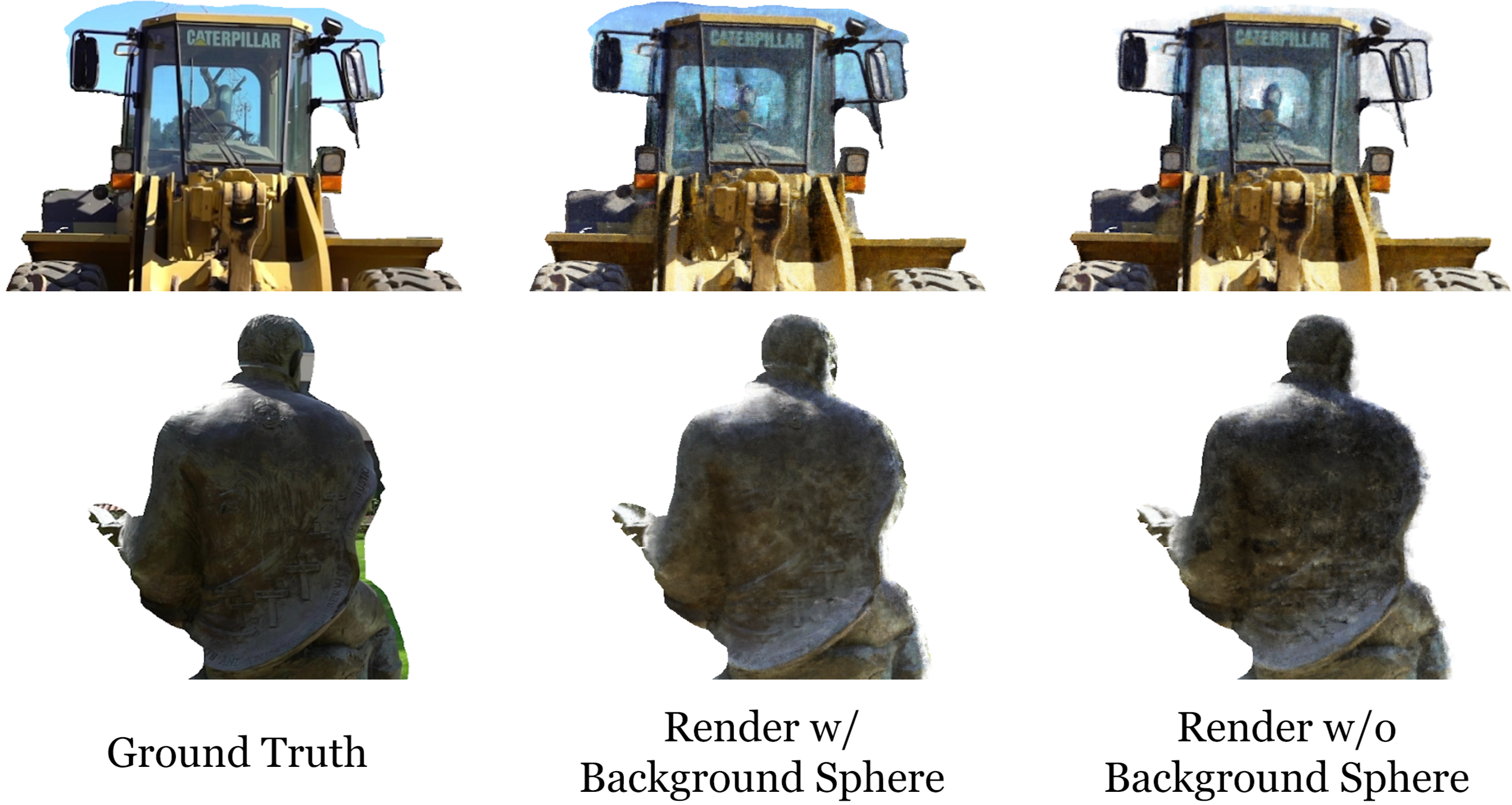}}
\caption{Modeling a background sphere helps \themodel to differentiate between the background and the foreground scene, thus preventing it from growing wrong points to model the background. Doing so potentially improves the robustness of the point cloud-guided NeRF model.} 
\vspace{-1em}
\label{fig:bgsph}
\end{center}
\end{figure}

Another limitation of the point cloud-guided NeRF models is their non-robustness against inaccurate background masks, as discussed in Section \ref{sec:tanks}. In the existing datasets, a mask is often given to distinguish the foreground from the background, so we only need to train NeRF on the foreground. However, if the mask is not precise, some regions of the background would be mistakenly included in the mask, as in the case of the Tanks and Temples dataset (\eg, the sky in the Caterpillar scene in Figure \ref{fig:bgsph}). Since a point cloud-guided NeRF model requires points to represent the entire scene including the background, it will try to grow the points in those background regions and use them to represent the background. On the other hand, the background regions can be far from the foreground, so the model will grow the points near the scene object rather than their true location. In normal situations, each point only represents a specific region of the scene. By contrast, for each of these mistakenly grown points, its view-dependent color is trained with different regions of the background which are inconsistent, thus resulting in abnormal rendering results in novel views.

To address this issue, we enhanced \themodel by modeling a {\em background sphere} that covers the entire scene. Therefore, the point cloud-guided NeRF model can directly represent the scene's background using the sphere, instead of growing new points. Such a strategy was introduced for the Tanks and Temples dataset in Figure~\ref{fig:highres-exp-tank}. Figure~\ref{fig:bgsph} further analyzes the impact of this background sphere, validating its effectiveness in approximately modeling background regions without growing wrong points. Further investigation is worthwhile in other strategies to tackle this issue.

\subsection{Environment Modeling in Shape Deformation Task}

Neither our work nor existing methods~\cite{nerfediting,deformingnerf,cagenerf} take into account the surrounding ambient environment when addressing the shape deformation task. These methods thus cannot assign different colors to the scene according to the changes in lighting conditions, as shown in the top of the shovel and the bent chimney in the Lego scene {and the shadow on the cushion in the Chair scene} (Figure~\ref{fig:highres-exp-all}). Fortunately, our \themodel, incorporating the Phong reflection's visual attributes (\eg, tint), enables fast fitting to the ambient environment through fine-tuning on the deformed scene (Section \ref{sec:ablation}).

\subsection{Evaluation for Scene Morphing Task} 

{\em Quantitatively} evaluating the visual realism of intermediate scenes during morphing is challenging, due to the lack of ground truth and associated metrics, as these scenes ``do not exist'' in the real world. Therefore, we rely mainly on visualizations for evaluation.

\begin{table*}[ht]
\centering

\scalebox{0.7}{\begin{tabular}{l|cccccccc|cccccccc}
 \hline\hline
 \multirow{2}{*}{Model} & \multicolumn{8}{c|}{Zero-Shot Inference} & \multicolumn{8}{c}{Fine-Tune for 1 Epoch} \\
 \cline{2-17}
& Chair & Hotdog & Lego & Drums & Ficus & Materials & Mic & Ship & Chair & Hotdog & Lego & Drums & Ficus & Materials & Mic & Ship  \\
\hline
\multicolumn{1}{c|}{} & \multicolumn{16}{c}{PSNR $\uparrow$} \\
 \hline
 DeformingNeRF~\cite{deformingnerf} & 18.84 & - & 13.10 &- &- &- &- &- &- &- &- &- &- &- &- &-  \\
 PointNeRF~\cite{pointnerf}  &22.21 &25.95 &24.56 &21.00 &24.24 &21.21 &26.77 &21.19 &30.11 &36.08 &31.45 &27.16 &31.48 &27.55 &34.34 &28.90 \\
 Naive Plotting  &24.91 &27.01 &25.64 &21.29 &26.22 &21.65 &27.63 &22.29 &32.01 &36.38 &31.72 &28.09 &33.21 &30.31 &35.15 &30.01 \\
 \hline
 \themodel w/o IST  &24.92 &27.02 &25.65 &21.29 &26.24 &21.64 &27.64 &22.28 &32.24 &36.69 &32.79 &28.30 &33.34 &30.40 &35.28 &30.08\\
 \themodel (Ours)  &\textbf{25.85} &\textbf{27.49} &\textbf{27.46} &\textbf{21.84} &\textbf{27.19} &\textbf{23.18} &\textbf{27.75} &\textbf{24.16} &\textbf{32.53} &\textbf{37.22} &\textbf{32.95} &\textbf{28.35} &\textbf{33.53} &\textbf{30.82} &\textbf{35.46} &\textbf{30.44}\\
 \hline

\multicolumn{1}{c|}{} & \multicolumn{16}{c}{SSIM $\uparrow$} \\
 \hline
 DeformingNeRF & 0.865 & - & 0.645 &- &- &- &- &- &- &- &- &- &- &- &- &-  \\
 PointNeRF  &0.910 &0.947 &0.933 &0.890 &0.915 &0.856 &0.945 &0.759 &0.972 &0.988 &0.977 &0.953 &0.973 &0.925 &0.981 &0.875 \\
 Naive Plotting  &0.950& 0.954& 0.934& 0.908& 0.953& 0.887& 0.964& 0.844& 0.984& 0.987& 0.977& 0.962& 0.985& 0.968& 0.988& 0.935 \\
 \hline
 \themodel w/o IST  &0.950 &0.954 &0.934 &0.908 &0.953 &0.887 &0.964 &0.845 &0.984 &0.988 &0.982 &\textbf{0.963} &0.985 &0.968 &0.988 &0.936\\
 \themodel (Ours)  &\textbf{0.963} &\textbf{0.960} &\textbf{0.966} &\textbf{0.916} &\textbf{0.960} &\textbf{0.909} &\textbf{0.966} &\textbf{0.875} &\textbf{0.986} &\textbf{0.989} &\textbf{0.983} &\textbf{0.963} &\textbf{0.986} &\textbf{0.971} &\textbf{0.989} &\textbf{0.939}\\
 \hline

\multicolumn{1}{c|}{} & \multicolumn{16}{c}{LPIPS AlexNet $\downarrow$} \\
 \hline
 DeformingNeRF & 0.071 & - & 0.297 &- &- &- &- &- &- &- &- &- &- &- &- &-  \\
 PointNeRF  &0.049& 0.080& 0.067& 0.122& 0.062& 0.099& 0.057& 0.139& 0.018& 0.033& 0.023& 0.077& 0.028& 0.067& 0.035& \textbf{0.073} \\
 Naive Plotting  &0.038& 0.063& 0.053& 0.088& 0.047& 0.098& 0.039& 0.159& 0.014& 0.023& 0.019& 0.047& 0.021& 0.042& 0.020& 0.081 \\
 \hline
 \themodel w/o IST  &0.037 &0.062 &0.052 &0.087 &0.046 &0.097 &0.039 &0.158 &0.013 &0.021 &0.015 &0.046 &0.020 &0.041 &\textbf{0.019} &0.080\\
 \themodel (Ours)  &\textbf{0.030} &\textbf{0.057} &\textbf{0.029} &\textbf{0.080} &\textbf{0.042} &\textbf{0.076} &\textbf{0.037} &\textbf{0.126} &\textbf{0.012} &\textbf{0.020} &\textbf{0.015} &\textbf{0.045} &\textbf{0.019} &\textbf{0.035} &\textbf{0.019} &0.075\\
 \hline

\multicolumn{1}{c|}{} & \multicolumn{16}{c}{LPIPS VGG $\downarrow$} \\
 \hline
 DeformingNeRF & 0.067 & - & 0.291 &- &- &- &- &- &- &- &- &- &- &- &- &-  \\
 PointNeRF  &0.047 &0.082 &0.060 &0.115 &0.070 &\textbf{0.091} &0.044 &\textbf{0.153} &\textbf{0.019} &0.055 &0.055 &0.086 &\textbf{0.039} &0.061 &0.029 &\textbf{0.090} \\
 Naive Plotting  &0.051 &0.080 &0.091 &0.094 &0.070 &0.109 &0.040 &0.183 &0.029 &0.050 &0.049 &0.068 &0.044 &0.065 &0.030 &0.129 \\
 \hline
 \themodel w/o IST  &0.051& 0.079& 0.088& 0.093& 0.069& 0.107& 0.040& 0.182& 0.027& 0.048& 0.043& 0.066& 0.042& 0.064& 0.029& 0.127\\
 \themodel (Ours)  &\textbf{0.041} &\textbf{0.074} &\textbf{0.057} &\textbf{0.088} &\textbf{0.067} &{0.095} &\textbf{0.038} &{0.163} &{0.026} &\textbf{0.045} &\textbf{0.042} &\textbf{0.065} &{0.042} &\textbf{0.058} &\textbf{0.028} &{0.121}\\

 \hline\hline
\end{tabular}}
\vspace{-3mm}
\caption{{{\bf Full comparison results of Table 1} in the main paper under all metrics. {\themodel} {\em significantly and consistently} outperforms PointNeRF and Naive Plotting on all deformed scenes of NeRF Synthetic {\em under all metrics}, in both zero-shot inference and fine-tuning settings. Our infinitesimal surface transformation (IST) effectively improves the results by correcting the view-dependent colors. With the precise point cloud generated by \themodel, even Naive Plotting consistently outperforms PointNeRF.} }
\vspace{-3mm}

\label{tab:exp-deform-complete}
\end{table*}

\begin{figure*}[t!]
\begin{center}
\centerline{\includegraphics[width=\linewidth]{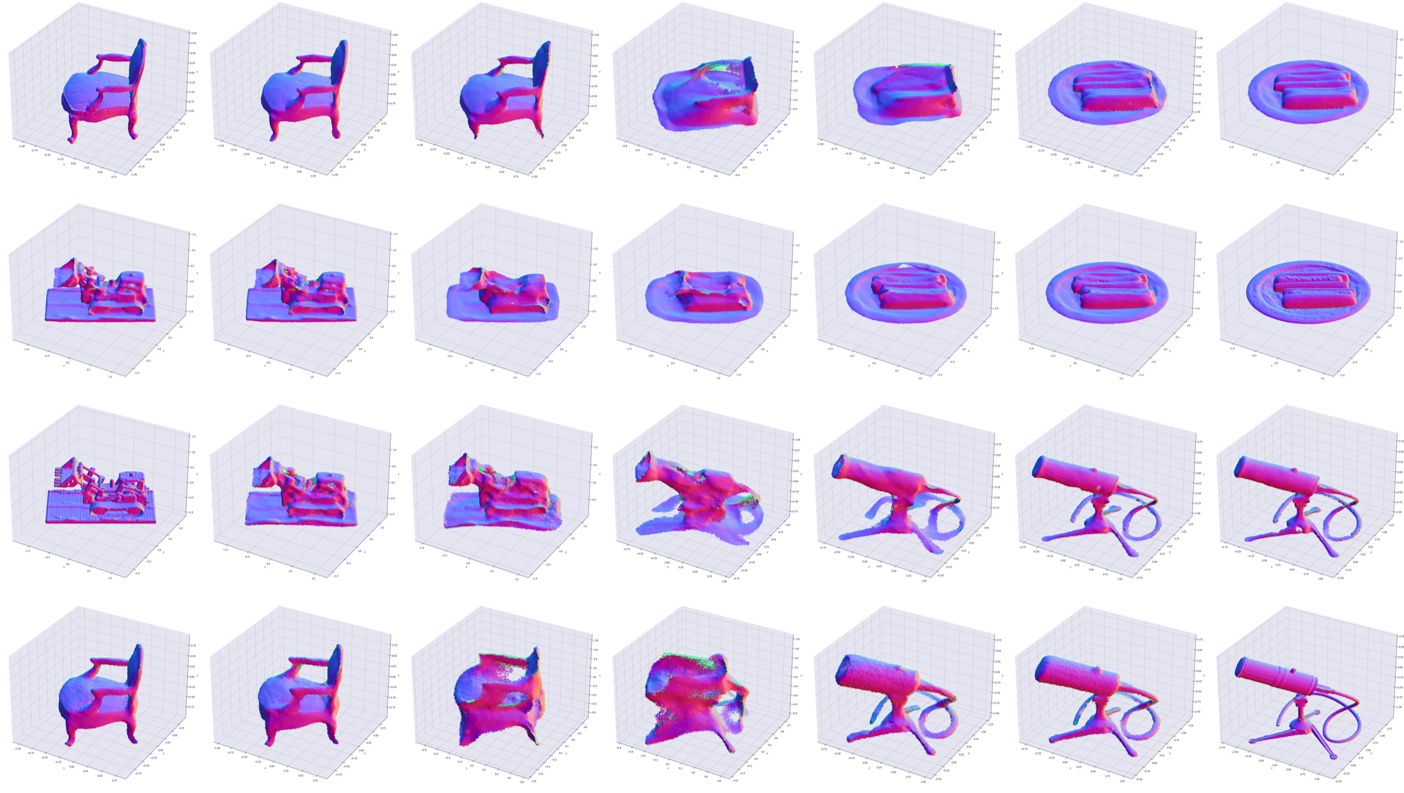}}
\caption{{\bf Intermediate point clouds for the scene morphing task corresponding to Figure 9} in the main paper, which are generated by the point cloud diffusion model~\cite{clouddiffusion}. } 
\label{fig:morph-cloud}
\end{center}
\end{figure*}
\begin{figure*}[t!]
\begin{center}
\centerline{\includegraphics[width=0.9\linewidth]{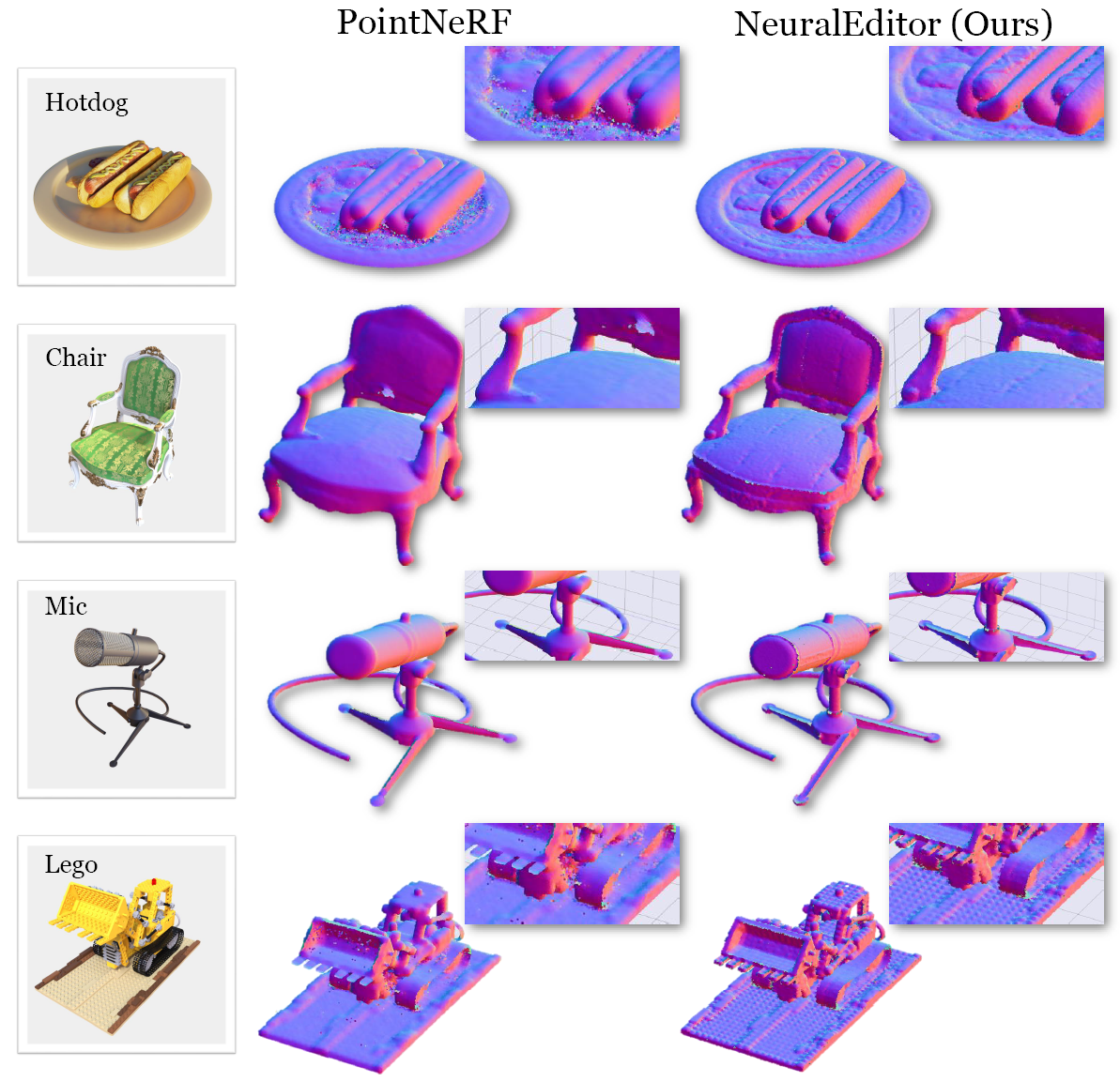}}
\caption{{\bf High-resolution version of Figure 6} in the main paper for more detailed visualization. \themodel generates much more precise point clouds than PointNeRF~\cite{pointnerf} in the four scenes of NeRF Synthetic~\cite{nerf}. The points are colored with their normal vectors.} 
\label{fig:highres-exp-cloud}
\end{center}
\end{figure*}
\vspace{-1em}
\begin{figure*}[t!]
\begin{center}
\centerline{\includegraphics[width=0.95\linewidth]{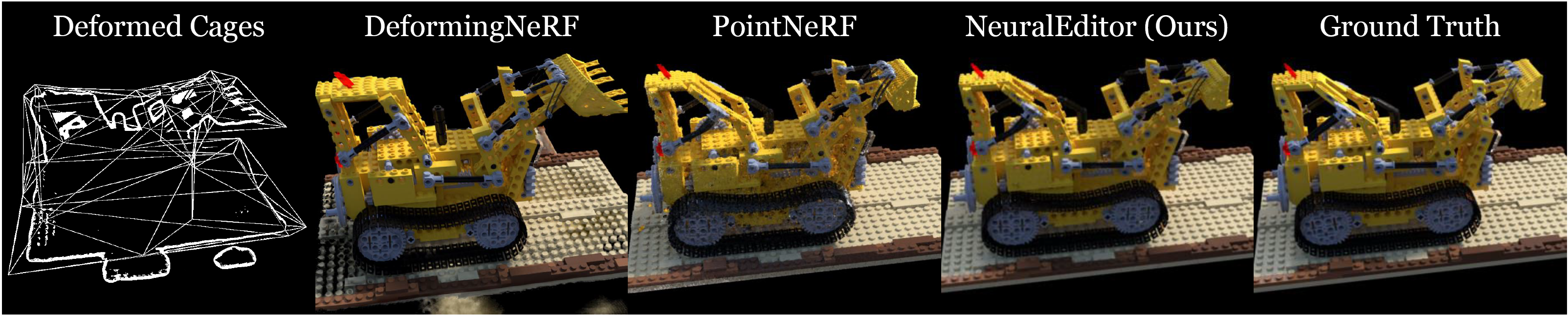}}
\caption{{\bf High-resolution version of Figure 7} in the main paper for more detailed visualization. With too coarse cages, Deforming-NeRF~\cite{deformingnerf} is unable to perform the deformation faithfully, leading to poor results.}
\label{fig:highres-exp-dnerf}
\end{center}
\end{figure*}
\begin{figure*}[t!]
\begin{center}
\centerline{\includegraphics[width=0.95\linewidth]{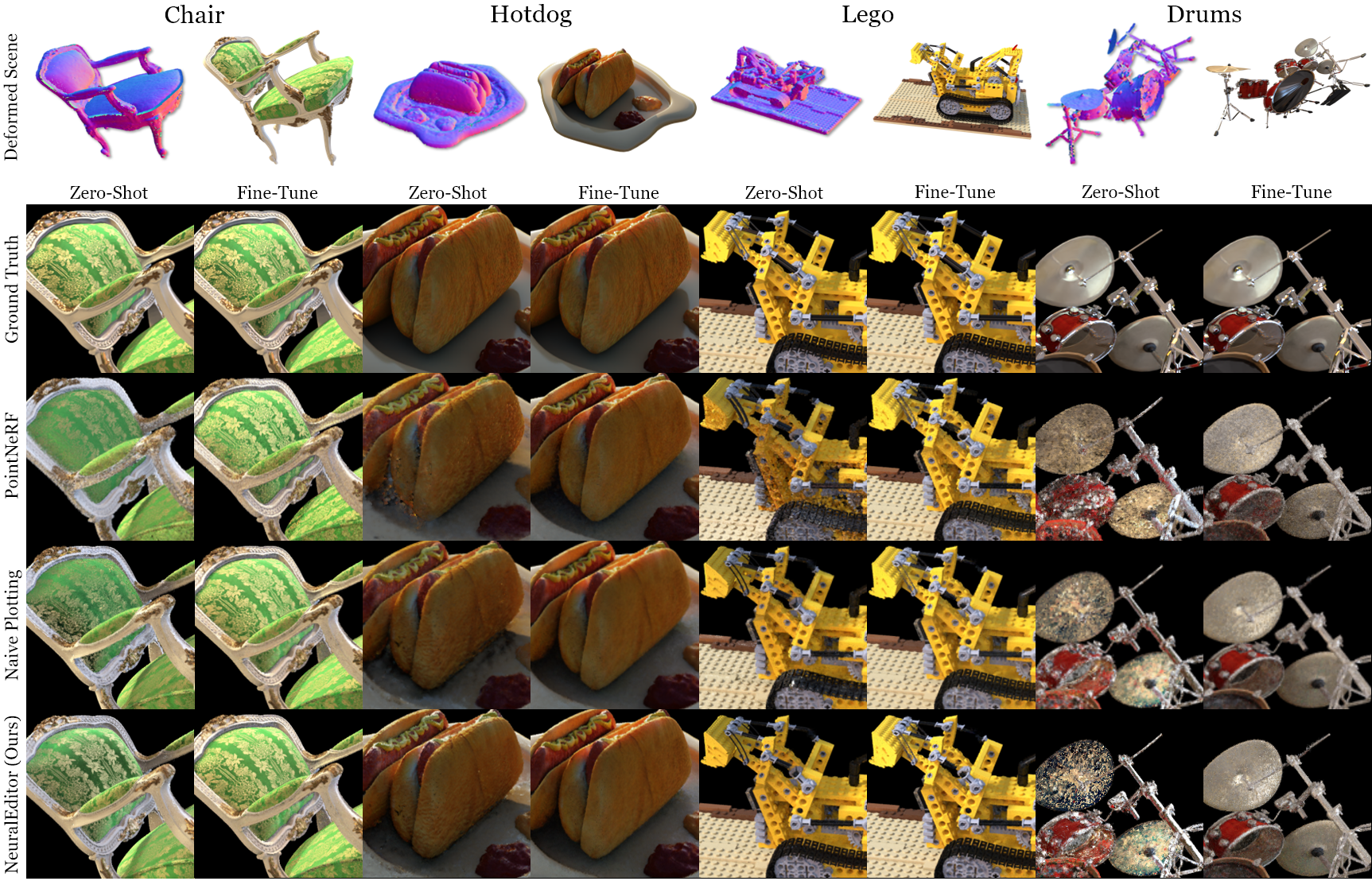}}
\centerline{\includegraphics[width=0.95\linewidth]{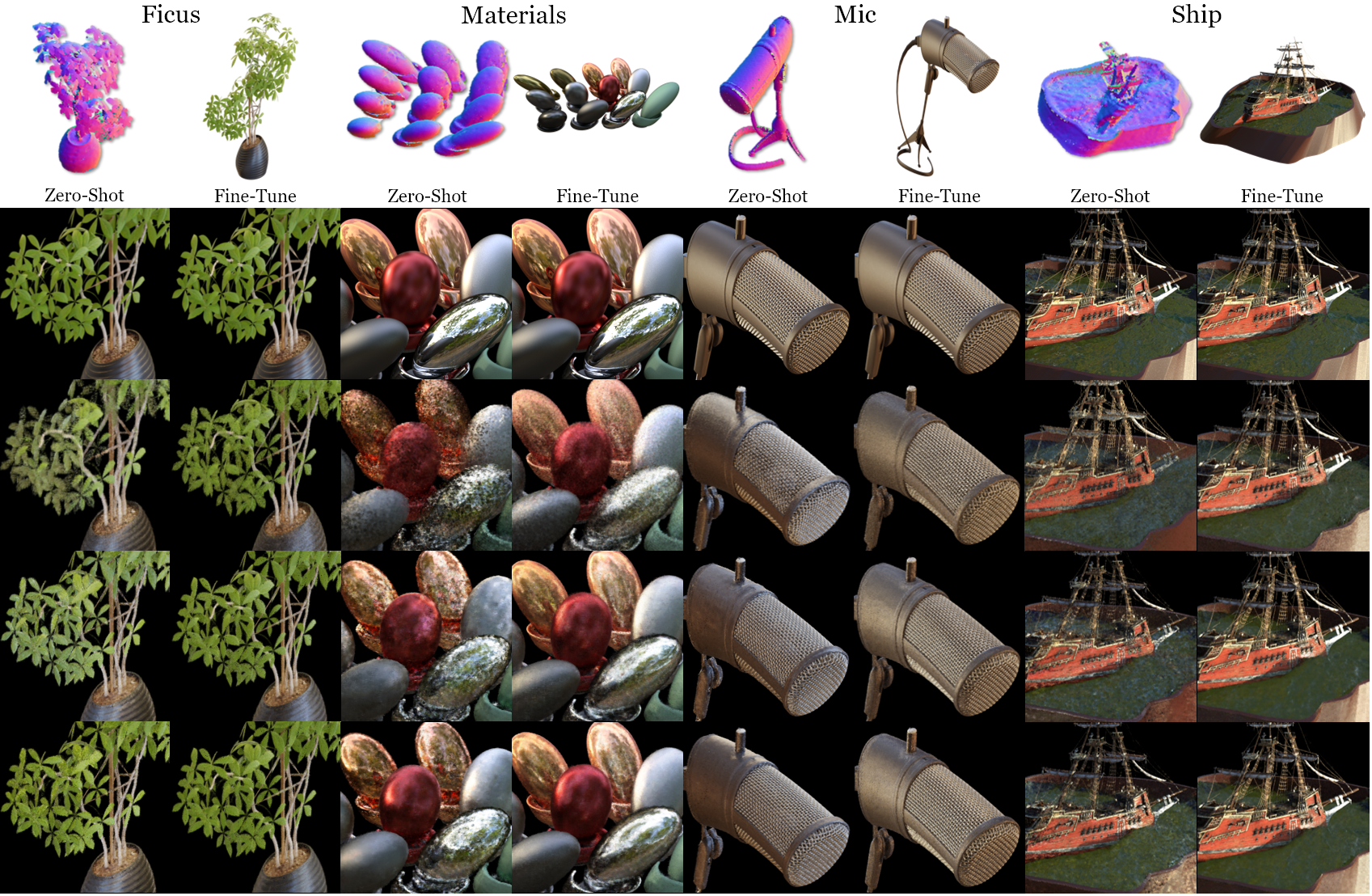}}
\caption{{\bf High-resolution version of Figure 8} in the main paper for more detailed visualization. {\themodel produces superior rendering results to PointNeRF, with significantly fewer artifacts in zero-shot inference. Fine-tuning further improves the {\em consistency of rendering with the ambient environment}. We use a black background for better visualization.} }
\label{fig:highres-exp-all}
\end{center}
\end{figure*}
\begin{figure*}[t!]
\centering
\centerline{\includegraphics[width=1\linewidth]{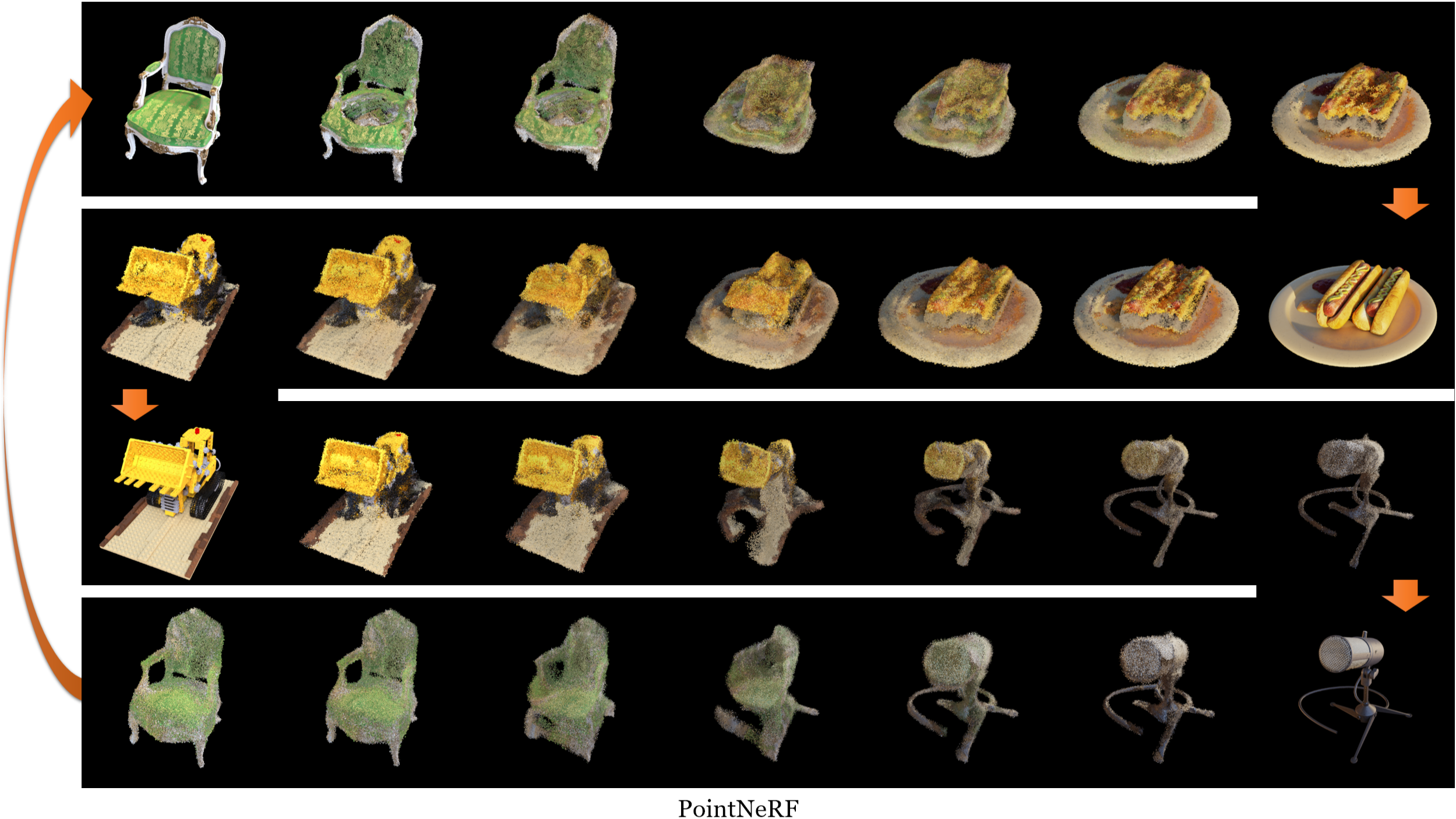}}
\centerline{\includegraphics[width=1\linewidth]{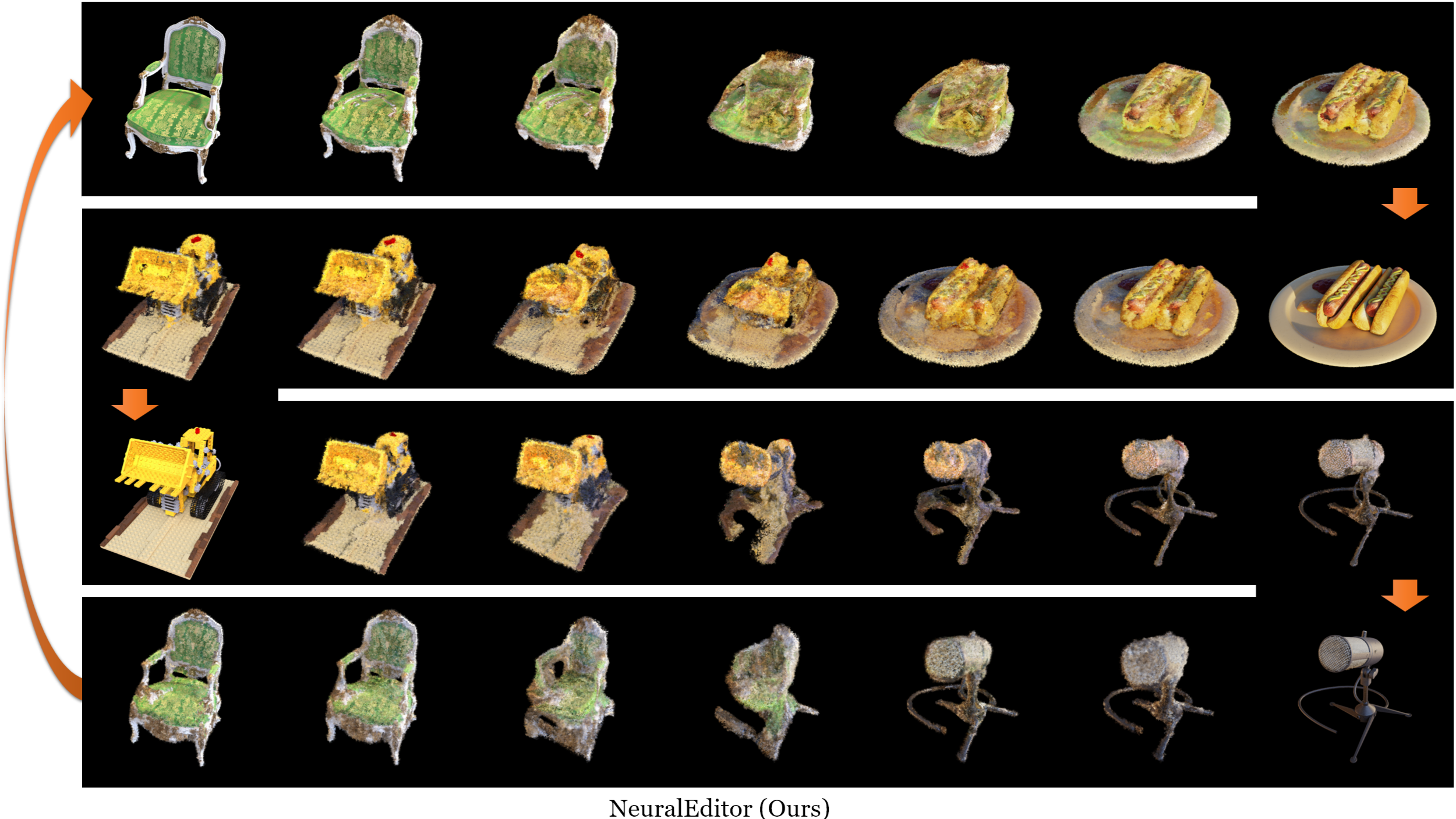}}
\caption{{\bf High-resolution, extended version of Figure 9} in the main paper for more detailed visualization. Our \themodel produces {\em smooth morphing} results between Chair, Hotdog, Lego, and Mic in the NeRF Synthetic dataset, while PointNeRF produces results with blurry textures, black shadows, and gloomy, non-smooth colors. The rendering results in the looped morphing process are arranged in the shape of the numerical digit ``3,'' indicated by the dividing lines and arrows. For the baseline PointNeRF, in the main paper we showed the morphing results between Chair and Hotdog due to limited space; here we include the full morphing results across all 4 scenes.}
\label{fig:highres-exp-morph}
\end{figure*}

%% file: main.bbl
\begin{thebibliography}{10}\itemsep=-1pt

\bibitem{dtu}
Henrik Aan{\ae}s, Rasmus~Ramsb{\o}l Jensen, George Vogiatzis, Engin Tola, and
  Anders~Bjorholm Dahl.
\newblock Large-scale data for multiple-view stereopsis.
\newblock {\em International Journal of Computer Vision}, 120(2):153--168,
  2016.

\bibitem{tradexp1}
Panos Achlioptas, Olga Diamanti, Ioannis Mitliagkas, and Leonidas~J. Guibas.
\newblock Learning representations and generative models for {3D} point clouds.
\newblock In {\em ICML}, 2018.

\bibitem{mipnerf}
Jonathan~T. Barron, Ben Mildenhall, Matthew Tancik, Peter Hedman, Ricardo
  Martin-Brualla, and Pratul~P. Srinivasan.
\newblock {Mip-NeRF}: A multiscale representation for anti-aliasing neural
  radiance fields.
\newblock In {\em ICCV}, 2021.

\bibitem{kdtree}
Jon~Louis Bentley.
\newblock Multidimensional binary search trees used for associative searching.
\newblock {\em Commun. ACM}, 18(9):509–517, 1975.

\bibitem{mvsnerf}
Anpei Chen, Zexiang Xu, Fuqiang Zhao, Xiaoshuai Zhang, Fanbo Xiang, Jingyi Yu,
  and Hao Su.
\newblock {MVSNeRF}: Fast generalizable radiance field reconstruction from
  multi-view stereo.
\newblock In {\em ICCV}, 2021.

\bibitem{tradimp4}
Zhiqin Chen and Hao Zhang.
\newblock Learning implicit fields for generative shape modeling.
\newblock In {\em CVPR}, 2019.

\bibitem{blender}
Blender~Online Community.
\newblock {\em Blender - a {3D} modelling and rendering package}.
\newblock Blender Foundation, Stichting Blender Foundation, Amsterdam, 2018.

\bibitem{nerfsurvey1}
Frank Dellaert and Lin Yen-Chen.
\newblock Neural volume rendering: {NeRF} and beyond.
\newblock {\em arXiv:2101.05204}, 2021.

\bibitem{nerfsurvey2}
Kyle Gao, Yina Gao, Hongjie He, Denning Lu, Linlin Xu, and Jonathan Li.
\newblock {NeRF}: Neural radiance field in {3D} vision, a comprehensive review.
\newblock {\em arXiv:2101.05204}, 2021.

\bibitem{nerfren}
Yuan-Chen Guo, Di Kang, Linchao Bao, Yu He, and Song-Hai Zhang.
\newblock {NeRFReN}: Neural radiance fields with reflections.
\newblock In {\em CVPR}, 2022.

\bibitem{tradexp3}
Peter Hedman, Tobias Ritschel, George Drettakis, and Gabriel Brostow.
\newblock Scalable inside-out image-based rendering.
\newblock {\em ACM Trans. Graph.}, 35(6):231:1--231:11, 2016.

\bibitem{deepmvs}
Po-Han Huang, Kevin Matzen, Johannes Kopf, Narendra Ahuja, and Jia-Bin Huang.
\newblock {DeepMVS}: Learning multi-view stereopsis.
\newblock In {\em CVPR}, 2018.

\bibitem{ske2}
Alec Jacobson, Ilya Baran, Ladislav Kavan, Jovan Popovi{\'{c}}, and Olga
  Sorkine.
\newblock Fast automatic skinning transformations.
\newblock {\em ACM Trans. Graph.}, 31(4), 2012.

\bibitem{cage2}
Tomas Jakab, Richard Tucker, Ameesh Makadia, Jiajun Wu, Noah Snavely, and
  Angjoo Kanazawa.
\newblock Keypointdeformer: Unsupervised {3D} keypoint discovery for shape
  control.
\newblock In {\em CVPR}, 2021.

\bibitem{tradexp5}
Mengqi Ji, Juergen Gall, Haitian Zheng, Yebin Liu, and Lu Fang.
\newblock Surfacenet: An end-to-end {3D} neural network for multiview
  stereopsis.
\newblock In {\em ICCV}, 2017.

\bibitem{cage0}
Tao Ju, Scott Schaefer, and Joe Warren.
\newblock Mean value coordinates for closed triangular meshes.
\newblock {\em ACM Trans. Graph.}, 24(3):561–566, 2005.

\bibitem{tank}
Arno Knapitsch, Jaesik Park, Qian-Yi Zhou, and Vladlen Koltun.
\newblock {Tanks and Temples}: Benchmarking large-scale scene reconstruction.
\newblock {\em ACM Trans. Graph.}, 36(4), 2017.

\bibitem{distillnerf}
Sosuke Kobayashi, Eiichi Matsumoto, and Vincent Sitzmann.
\newblock Decomposing {NeRF} for editing via feature field distillation.
\newblock In {\em NeurIPS}, 2022.

\bibitem{tradimp3}
Marc Levoy and Pat Hanrahan.
\newblock Light field rendering.
\newblock In {\em SIGGRAPH}, 1996.

\bibitem{tradexp6}
Fayao Liu, Chunhua Shen, Guosheng Lin, and Ian Reid.
\newblock Learning depth from single monocular images using deep convolutional
  neural fields.
\newblock {\em IEEE Transactions on Pattern Analysis and Machine Intelligence},
  38(10):2024--2039, 2015.

\bibitem{neuralsparse}
Lingjie Liu, Jiatao Gu, Kyaw~Zaw Lin, Tat-Seng Chua, and Christian Theobalt.
\newblock Neural sparse voxel fields.
\newblock In {\em NeurIPS}, 2020.

\bibitem{editnerf}
Steven Liu, Xiuming Zhang, Zhoutong Zhang, Richard Zhang, Junyan Zhu, and
  Bryan~C. Russell.
\newblock Editing conditional radiance fields.
\newblock In {\em ICCV}, 2021.

\bibitem{marchingcubes}
William~E. Lorensen and Harvey~E. Cline.
\newblock Marching cubes: A high resolution {3D} surface construction
  algorithm.
\newblock In {\em SIGGRAPH}, 1987.

\bibitem{clouddiffusion}
Shitong Luo and Wei Hu.
\newblock Diffusion probabilistic models for {3D} point cloud generation.
\newblock In {\em CVPR}, 2021.

\bibitem{ske1}
Bruce Merry, Patrick Marais, and James Gain.
\newblock Animation space: A truly linear framework for character animation.
\newblock {\em ACM Trans. Graph.}, 25(4):1400–1423, 2006.

\bibitem{tradimp2}
Lars Mescheder, Michael Oechsle, Michael Niemeyer, Sebastian Nowozin, and
  Andreas Geiger.
\newblock Occupancy networks: Learning {3D} reconstruction in function space.
\newblock In {\em CVPR}, 2019.

\bibitem{nerf}
Ben Mildenhall, Pratul~P. Srinivasan, Matthew Tancik, Jonathan~T. Barron, Ravi
  Ramamoorthi, and Ren Ng.
\newblock {NeRF}: Representing scenes as neural radiance fields for view
  synthesis.
\newblock In {\em ECCV}, 2020.

\bibitem{tradimp1}
Michael Niemeyer, Lars Mescheder, Michael Oechsle, and Andreas Geiger.
\newblock Differentiable volumetric rendering: Learning implicit {3D}
  representations without {3D} supervision.
\newblock In {\em CVPR}, 2020.

\bibitem{cagesurvey}
Jesús Nieto and Toni Susin.
\newblock Cage based deformations: A survey.
\newblock {\em Lecture Notes in Computational Vision and Biomechanics},
  7:75--99, 2013.

\bibitem{cagenerf}
Yicong Peng, Yichao Yan, Shengqi Liu, Yuhao Cheng, Shanyan Guan, Bowen Pan,
  Guangtao Zhai, and Xiaokang Yang.
\newblock Cage{NeRF}: Cage-based neural radiance field for generalized {3D}
  deformation and animation.
\newblock In {\em NeurIPS}, 2022.

\bibitem{phong}
Bui~Tuong Phong.
\newblock Illumination for computer generated pictures.
\newblock {\em Commun. ACM}, 18(6):311–317, 1975.

\bibitem{neuphysics}
Yi-Ling Qiao, Alexander Gao, and Ming~C. Lin.
\newblock {NeuPhysics}: Editable neural geometry and physics from monocular
  videos.
\newblock In {\em NeurIPS}, 2022.

\bibitem{plenoxels}
{Sara Fridovich-Keil and Alex Yu}, Matthew Tancik, Qinhong Chen, Benjamin
  Recht, and Angjoo Kanazawa.
\newblock {Plenoxels}: Radiance fields without neural networks.
\newblock In {\em CVPR}, 2022.

\bibitem{tradexp0}
Noah Snavely, Steven~M. Seitz, and Richard Szeliski.
\newblock Photo tourism: Exploring photo collections in {3D}.
\newblock {\em ACM Trans. Graph.}, 25(3):835–846, 2006.

\bibitem{ccnerf}
Jiaxiang Tang, Xiaokang Chen, Jingbo Wang, and Gang Zeng.
\newblock Compressible-composable {NeRF} via rank-residual decomposition.
\newblock In {\em NeurIPS}, 2022.

\bibitem{cage4}
Jean-Marc Thiery, Julien Tierny, and Tamy Boubekeur.
\newblock Jacobians and hessians of mean value coordinates for closed
  triangular meshes.
\newblock {\em The Visual Computer}, 30(9):981--995, 2014.

\bibitem{refnerf}
Dor Verbin, Peter Hedman, Ben Mildenhall, Todd Zickler, Jonathan~T. Barron, and
  Pratul~P. Srinivasan.
\newblock {Ref-NeRF}: Structured view-dependent appearance for neural radiance
  fields.
\newblock In {\em CVPR}, 2022.

\bibitem{tradexp2}
Nanyang Wang, Yinda Zhang, Zhuwen Li, Yanwei Fu, Wei Liu, and Yu-Gang Jiang.
\newblock {Pixel2Mesh}: Generating {3D} mesh models from single {RGB} images.
\newblock In {\em ECCV}, 2018.

\bibitem{ibrnet}
Qianqian Wang, Zhicheng Wang, Kyle Genova, Pratul Srinivasan, Howard Zhou,
  Jonathan~T. Barron, Ricardo Martin-Brualla, Noah Snavely, and Thomas
  Funkhouser.
\newblock {IBRNet}: Learning multi-view image-based rendering.
\newblock In {\em CVPR}, 2021.

\bibitem{nex}
Suttisak Wizadwongsa, Pakkapon Phongthawee, Jiraphon Yenphraphai, and Supasorn
  Suwajanakorn.
\newblock {NeX}: Real-time view synthesis with neural basis expansion.
\newblock In {\em CVPR}, 2021.

\bibitem{diver}
Liwen Wu, {Jae Yong} Lee, Anand Bhattad, Yu-Xiong Wang, and David Forsyth.
\newblock {DIVeR}: Real-time and accurate neural radiance fields with
  deterministic integration for volume rendering.
\newblock In {\em CVPR}, 2022.

\bibitem{insp}
Dejia Xu, Peihao Wang, Yifan Jiang, Zhiwen Fan, and Zhangyang Wang.
\newblock Signal processing for implicit neural representations.
\newblock In {\em NeurIPS}, 2022.

\bibitem{pointnerf}
Qiangeng Xu, Zexiang Xu, Julien Philip, Sai Bi, Zhixin Shu, Kalyan Sunkavalli,
  and Ulrich Neumann.
\newblock {Point-NeRF}: Point-based neural radiance fields.
\newblock In {\em CVPR}, 2021.

\bibitem{deformingnerf}
Tianhan Xu and Tatsuya Harada.
\newblock Deforming radiance fields with cages.
\newblock In {\em ECCV}, 2022.

\bibitem{objectnerf}
Bangbang Yang, Yinda Zhang, Yinghao Xu, Yijin Li, Han Zhou, Hujun Bao, Guofeng
  Zhang, and Zhaopeng Cui.
\newblock Learning object-compositional neural radiance field for editable
  scene rendering.
\newblock In {\em ICCV}, 2021.

\bibitem{mvsnet}
Yao Yao, Zixin Luo, Shiwei Li, Tian Fang, and Long Quan.
\newblock {MVSNet}: Depth inference for unstructured multi-view stereo.
\newblock In {\em ECCV}, 2018.

\bibitem{cage1}
Wang Yifan, Noam Aigerman, Vladimir~G Kim, Siddhartha Chaudhuri, and Olga
  Sorkine-Hornung.
\newblock Neural cages for detail-preserving {3D} deformations.
\newblock In {\em CVPR}, 2020.

\bibitem{plenoct}
Alex Yu, Ruilong Li, Matthew Tancik, Hao Li, Ren Ng, and Angjoo Kanazawa.
\newblock {PlenOctrees} for real-time rendering of neural radiance fields.
\newblock In {\em ICCV}, 2021.

\bibitem{pixelnerf}
Alex Yu, Vickie Ye, Matthew Tancik, and Angjoo Kanazawa.
\newblock {pixelNeRF}: Neural radiance fields from one or few images.
\newblock In {\em CVPR}, 2021.

\bibitem{tradeditsurvey}
Yu-Jie Yuan, Yu-Kun Lai, Tong Wu, Lin Gao, and Ligang Liu.
\newblock A revisit of shape editing techniques: From the geometric to the
  neural viewpoint.
\newblock {\em Journal of Computer Science and Technology}, 36(3):520--554,
  2021.

\bibitem{nerfediting}
Yu-Jie Yuan, Yang tian Sun, Yu-Kun Lai, Yuewen Ma, Rongfei Jia, and Lin Gao.
\newblock {NeRF-Editing}: Geometry editing of neural radiance fields.
\newblock In {\em CVPR}, 2022.

\bibitem{cage3}
Yuzhe Zhang, Jianmin Zheng, and Yiyu Cai.
\newblock Proxy-driven free-form deformation by topology-adjustable control
  lattice.
\newblock {\em Computers \& Graphics}, 89:167--177, 2020.

\bibitem{open3d}
Qian-Yi Zhou, Jaesik Park, and Vladlen Koltun.
\newblock {Open3D}: {A} modern library for {3D} data processing.
\newblock {\em arXiv:1801.09847}, 2018.

\end{thebibliography}
